\documentclass{article}

\usepackage[preprint]{corl_2026} 
\usepackage{amsmath}
\usepackage{array}
\usepackage{xcolor}
\usepackage{makecell}
\usepackage{wrapfig}
\usepackage{booktabs}
\usepackage{graphicx}
\usepackage[table]{xcolor}
\usepackage{adjustbox}
\usepackage{enumitem}
\usepackage{multirow}

\newcommand{\occclass}[2]{%
  \rotatebox{90}{%
    \textcolor{#1}{\rule{1.25ex}{1.25ex}}%
    \hspace{0.45em}#2%
  }%
}

\usepackage{amssymb}
\newcommand{\cmark}{\checkmark}
\newcommand{\xmark}{--}

\title{VISA: VLM-Guided Instance Semantic Auditing for 3D Occupancy World Models}

\author{
  Ruiqi Xian\\
  University of Maryland College Park, USA 
  \AND
  Yuehan Xian \\
  Independent Researcher, USA \\
  \AND
  Jing Liang \\
  Standford University, USA \\
  \And
  Tony (Xuewei ) Qi \\
  Independent Researcher, USA \\
  \And
  Dinesh Manocha \\
  University of Maryland College Park, USA  \\
}

\begin{document}
\maketitle


\begin{abstract}
Semantic 3D occupancy provides a voxelized world state for autonomous driving and robot decision making, but object and rare-class errors can affect free-space interpretation, collision checking, and temporal state propagation. We show that a common VLM strategy, aligning 3D voxel or object features with crop-caption embeddings, improves text-space similarity without reliably improving closed-set occupancy mIoU. Motivated by this mismatch, we propose \textbf{VISA}, a training-time semantic auditing approach for existing occupancy world models. VISA queries an offline VLM on a representative crop of each physical object instance, obtains a structured audit with class hypotheses, plausible confusions, reliability, attributes, and evidence, and propagates it along the object track. The audit is grounded to matched 3D object voxels and distilled into semantic logits through reliability-weighted taxonomy, attribute-factor, and scene-level audit graph losses, while inference remains unchanged and requires no VLM. On nuScenes, averaged across three runs, VISA improves OccWorld from $19.06$ to $20.05$ mIoU and GaussianWorld from $21.36$ to $21.91$ mIoU; on GaussianWorld, object mIoU improves from $18.18$ to $19.16$ and rare-class mIoU from $15.60$ to $16.79$. These results suggest that VLMs are better suited to closed-set occupancy as reliability-aware semantic auditors than as generic caption-embedding targets.
\end{abstract}


\section{Introduction}

3D semantic occupancy prediction is becoming a central representation for autonomous driving perception, robot decision making, and occupancy-based world modeling~\cite{song2017sscnet,caesar2020nuscenes,huang2023tpvformer,wei2023surroundocc,tian2023occ3d,zheng2023occworld}. It represents a scene as a dense voxelized world state, where each voxel stores occupancy and semantic predictions for foreground objects, background regions, and free space~\cite{song2017sscnet,cao2022monoscene,li2023voxformer,wei2023surroundocc,tian2023occ3d}. Unlike object-centric perception, which often relies on sparse boxes or bird's-eye-view abstractions~\cite{li2022bevformer}, semantic occupancy provides a unified spatial representation that can support planning, collision checking, map updating, and temporal state propagation~\cite{ha2018worldmodels,zheng2023occworld,min2024driveworld,zuo2025gaussianworld}. Semantic reliability is therefore important: confusing traffic cones with barriers, trucks with buses or trailers, or pedestrians and motorcycles with background regions can change the free-space and obstacle semantics used by downstream autonomy modules. Although recent occupancy models have made strong progress in reconstructing occupied space from camera inputs~\cite{li2023voxformer,huang2023tpvformer,wei2023surroundocc}, learning semantically reliable world states remains challenging due to visually similar categories, long-tail objects, occlusion, and the fine-grained closed-set taxonomy used in occupancy benchmarks~\cite{caesar2020nuscenes,tian2023occ3d,wei2023surroundocc}.

Vision-language models (VLMs) offer semantic knowledge through transferable visual representations and image-language reasoning~\cite{radford2021clip,zhai2023siglip,li2023blip2,liu2023llava,bai2023qwenvl,wang2024qwen2vl}. This has motivated 3D scene understanding methods based on CLIP-style distillation, open-vocabulary feature alignment, and region-level language supervision~\cite{chen2023clip2scene,peng2023openscene,ding2022pla,yang2023regionplc}. A natural adaptation to semantic occupancy is to caption object crops and align 3D voxel or object features with the corresponding language embeddings~\cite{radford2021clip,zhai2023siglip,ding2022pla,chen2023clip2scene}. However, caption embeddings are open-vocabulary and instance-specific, whereas occupancy evaluation is closed-set, taxonomy-specific, and voxel-level~\cite{huang2023tpvformer,wei2023surroundocc,tian2023occ3d}. Object crops also cover sparse object regions, while occupancy models predict dense foreground, background, and free-space semantics across the full 3D volume~\cite{li2023voxformer,huang2023tpvformer,wei2023surroundocc,tian2023occ3d}. We verify this mismatch with a diagnostic ablation: instance-text alignment improves its auxiliary text-space objective, but does not reliably improve semantic occupancy mIoU. This motivates using VLMs as offline visual auditors rather than embedding targets~\cite{radford2021clip,zhai2023siglip,li2023blip2,liu2023llava,bai2023qwenvl,wang2024qwen2vl,chen2024internvl,li2024llavanext}. A crop-level audit can provide occupancy-aligned cues unavailable from a free-form caption: a closed-set class hypothesis, plausible visual confusions, attributes, reliability, and evidence. For occupancy training, the useful VLM signal should therefore be taxonomy-aware, instance-specific, reliability-aware, and grounded to voxel-level semantic decisions rather than preserved as a generic open-vocabulary embedding.

We propose \textbf{VISA}, a training-time semantic auditing approach for existing occupancy world models. VISA does not introduce a new occupancy architecture or change inference. It audits representative object crops with an offline VLM, converts the responses into structured audit tuples, and grounds them only to matched 3D object voxels of the same physical instance. During training, these audits supervise voxel semantic logits through reliability-weighted taxonomy, attribute-factor, and scene-level audit graph losses. At inference, the model uses the same camera inputs and occupancy world-model architecture as the baseline, with no VLM calls, text encoder, crop extraction, or additional modules. Compared with prior vision-language supervision for 3D perception, VISA differs in both target space and grounding. Existing methods transfer CLIP/VLM knowledge through open-vocabulary feature alignment, region-level language supervision, or voxel-space embedding and language-prior distillation~\cite{chen2023clip2scene,peng2023openscene,ding2022pla,yang2023regionplc,tan2023ovo,vobecky2024pop3d,zhang2025clip,boeder2024langocc,yu2024language,zheng2024veon,feng2024vipocc,doruk2026vlmfusion}. These approaches are useful for open-vocabulary recognition or vocabulary expansion, but closed-set occupancy is evaluated by voxel-level logits under a fixed driving taxonomy~\cite{huang2023tpvformer,wei2023surroundocc,tian2023occ3d}. VISA instead converts crop-level VLM judgments into reliability-aware taxonomy and attribute supervision grounded to matched object voxels. Additional related work is discussed in Appendix~\ref{app:related}. Our contributions are:
\begin{enumerate}[leftmargin=*, itemsep=1pt, topsep=2pt, parsep=0pt, partopsep=0pt]
\item We diagnose a task-specific failure mode of generic VLM supervision for occupancy: open-vocabulary caption/feature alignment can optimize text-space objectives while remaining weakly coupled to closed-set voxel-level semantic prediction.

\item We introduce \textbf{VISA}, a training-time semantic auditing approach that converts VLM crop understanding into reliability-aware closed-set taxonomy and visual-factor supervision, replacing generic caption embeddings with occupancy-compatible audit signals.

\item We propose audit-to-occupancy grounding for physical object instances: audits are associated with the same object over time, applied only to matched 3D object voxels, and further organized by a scene-level audit graph over co-occurring objects.

\item We validate VISA as training-time supervision for existing occupancy world models. Averaged across three runs, VISA improves semantic mIoU by $+0.99$ on OccWorld and $+0.55$ on GaussianWorld, with especially strong object/rare-class gains of $+1.40/+1.42$ on OccWorld and $+0.98/+1.19$ on GaussianWorld, and no inference-time VLM cost.
\end{enumerate}
\section{Problem Setup}
\label{sec:problem}
\begin{figure*}[t]
\centering
\includegraphics[width=\textwidth]{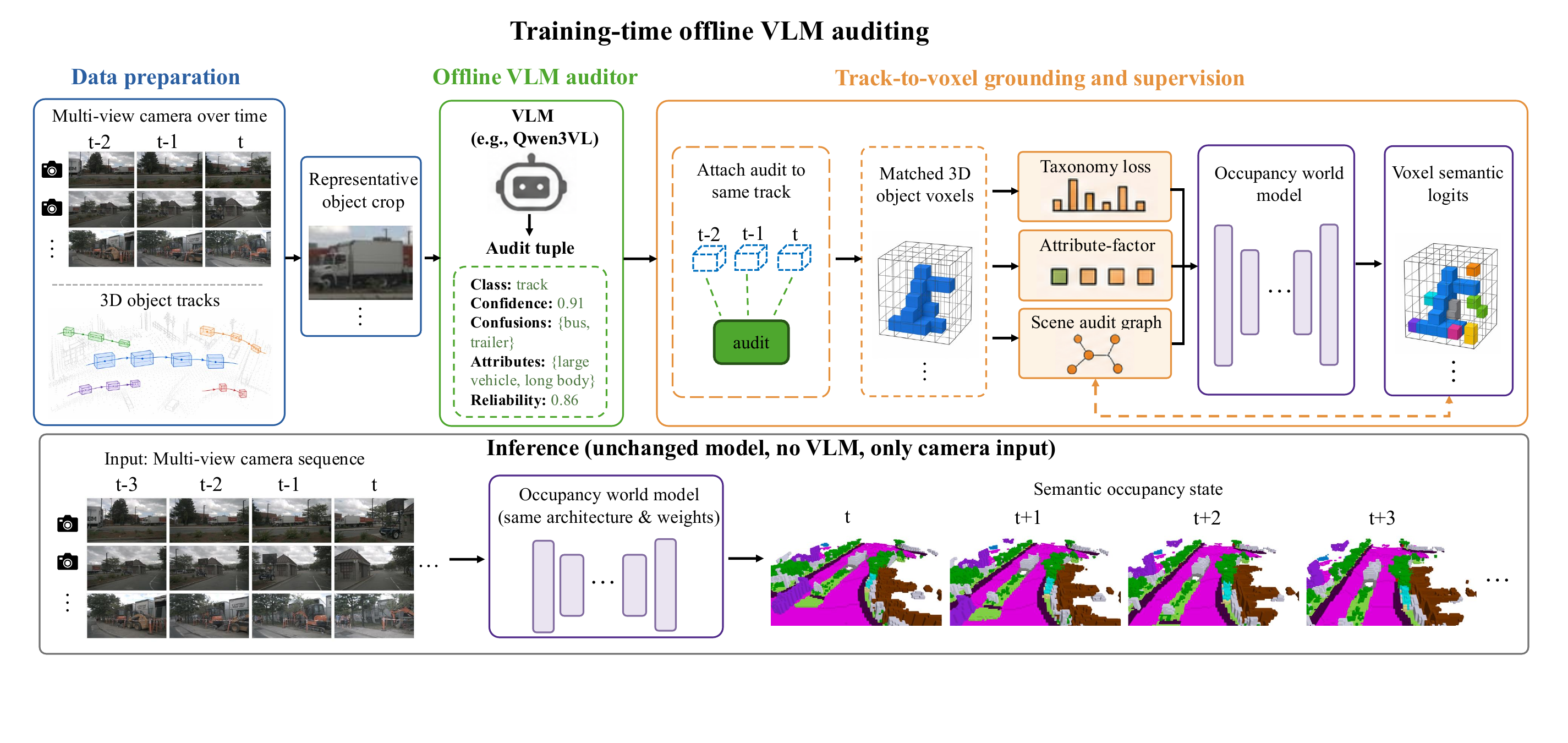}
\vspace{-4em}
\caption{
Overview of \textbf{VISA}. During training, object tracks are converted into representative crops and audited by an offline VLM to obtain closed-set class hypotheses, plausible confusions, attributes, reliability, and evidence. The audit is propagated to 3D boxes of the same physical instance, grounded to matched object voxels, and distilled into semantic occupancy logits through taxonomy, attribute, and scene-level audit graph supervision. At inference, the original occupancy world model is unchanged and requires no VLM, text encoder, crop extraction, or object-track input.
}
\label{fig:visa_method}
\vspace{-1em}
\end{figure*}

We study how offline VLM audits can provide training-time supervision for semantic 3D occupancy world models. The occupancy model predicts dense closed-set semantic logits over the full 3D scene, while the VLM observes sparse object-centric image crops and produces instance-level semantic judgments. The goal is to convert these crop-level audits into voxel-compatible supervision during training, without changing the inference-time occupancy model.

\textbf{Semantic 3D Occupancy World Model:}
The occupancy model $f_{\theta}$ takes a single frame or a sequence of camera observations, $\mathbf{I}_{1:t}$, and predicts dense semantic occupancy logits $\mathbf{Z}_t = f_{\theta}(\mathbf{I}_{1:t})$, where $\mathbf{Z}_t \in \mathbb{R}^{H \times W \times D \times C}$. Each voxel $v$ at time $t$ has a class-logit vector $\mathbf{z}_{v,t}\in\mathbb{R}^{C}$, where $C$ includes object classes, stuff classes, and the empty class. Given dense voxel labels $\mathbf{Y}_t$, standard training uses cross-entropy and Lovasz losses:
\begin{equation}
    \mathcal{L}_{\mathrm{occ}}
    =
    \mathcal{L}_{\mathrm{ce}}(\mathbf{Z}_t,\mathbf{Y}_t)
    +
    \mathcal{L}_{\mathrm{lovasz}}(\mathbf{Z}_t,\mathbf{Y}_t).
\end{equation}
VISA does not assume a particular internal state representation or temporal update rule; it only adds training-time supervision to the semantic logits.

\textbf{Offline VLM Audits and Object Voxels:}
During training, we additionally use object-level regions/tracks and offline VLM audits. For each physical object instance $i$, a representative image crop is queried with a closed-set taxonomy and visual-factor prompt. The VLM returns a JSON audit and we parse it as an audit tuple: 
\begin{equation}
    \mathcal{A}_i =
    \left(
    c_i^{\mathrm{vlm}},
    q_i,
    \mathcal{C}_i,
    \mathbf{u}_i,
    r_i,
    e_i
    \right),
    \label{eq:audit_tuple}
\end{equation}
where $c_i^{\mathrm{vlm}}$ is a closed-set class hypothesis, $q_i\in[0,1]$ is the VLM-reported confidence for that hypothesis, $\mathcal{C}_i$ is a set of plausible visual confusions with VLM-reported confidence scores, $\mathbf{u}_i\in\{0,1\}^{A}$ is a vector of occupancy-relevant visual factors, $r_i\in[0,1]$ is the visual reliability of the crop, and $e_i$ is a short evidence string used for inspection. These audits are sparse and object-centric: they describe visible object crops, not the full occupancy grid or stuff regions.

To use an audit for occupancy training, it must be grounded to the corresponding 3D object region. Let $\mathbf{b}_{i,t}$ be the 3D box of instance $i$ at time $t$, and let $y_i^{\mathrm{obj}}$ be its object class. Using the dense voxel label $y_{v,t}$ from $\mathbf{Y}_t$, we define the matched object-voxel set as
\begin{equation}
    \mathcal{V}_{i,t}
    =
    \{v \mid \mathbf{x}_{v,t} \in \mathbf{b}_{i,t},\; y_{v,t}=y_i^{\mathrm{obj}}\}.
    \label{eq:object_voxel_set}
\end{equation}
This training-time restriction prevents a crop-level VLM judgment from supervising nearby background, free space, or overlapping objects.

\textbf{Learning Objective:}
Given dense occupancy labels, object tracks, and offline audits, we train the occupancy model with the standard occupancy loss plus a VISA supervision term:
\begin{equation}
    \min_{\theta}
    \mathcal{L}_{\mathrm{occ}}(\theta)
    +
    \mathcal{L}_{\mathrm{VISA}}(\theta;\{\mathcal{A}_i,\mathcal{V}_{i,t}\}_{i,t}),
    \label{eq:objective_loss}
\end{equation}
where $i$ indexes audited object instances and $t$ indexes frames in which the corresponding instance has a 3D box. The role of $\mathcal{L}_{\mathrm{VISA}}$ is to convert instance-level VLM judgments into closed-set, voxel-grounded semantic supervision. Section~\ref{sec:method} instantiates this term with reliability-weighted taxonomy distillation, attribute-factor supervision, and scene-level audit graph regularization.

\begin{wrapfigure}{r}{0.56\linewidth}
    \vspace{-4.5em}
    \centering
    \includegraphics[width=\linewidth]{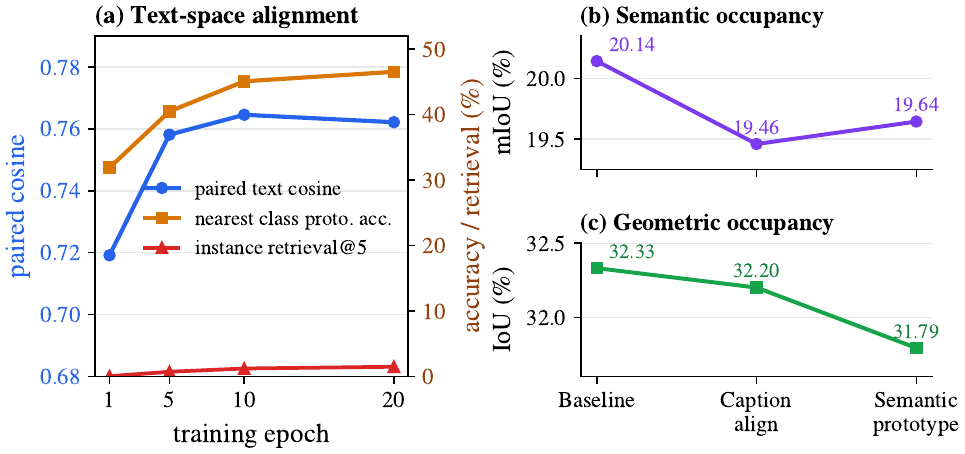}
    \vspace{-2em}
\caption{
Diagnostic of generic language alignment for occupancy.
Left: paired text cosine measures projected 3D feature--caption similarity, nearest class-prototype accuracy uses closed-set class text prototypes, and instance retrieval@5 retrieves the paired caption from the caption bank.
Right: caption alignment uses crop captions, while semantic prototype uses class text prototypes; both remain below the occupancy baseline.
}
    \label{fig:caption_alignment_diagnostic}
    \vspace{-1em}
\end{wrapfigure}
\section{VISA}
\label{sec:method}

VISA is a training-time semantic auditing approach for existing semantic 3D occupancy world models. Given offline object crops, VISA parses VLM responses into structured audit tuples, associates each audit with the same physical object track across time, grounds it to matched 3D object voxels, and distills it into voxel semantic logits. The VLM is used only offline; at inference, the underlying occupancy world model is unchanged. VISA consists of four components: (i) offline instance audit generation, (ii) track-to-voxel grounding, (iii) reliability-weighted taxonomy and attribute-factor distillation, and (iv) scene-level audit graph regularization over co-occurring audited objects.

\textbf{From Caption Alignment to Semantic Auditing:}
A common way to use offline VLM supervision is to caption each object crop, encode the caption with a text encoder, and align the corresponding 3D instance feature with the text embedding. Let $\mathbf{h}_{i,t}$ be the pooled 3D feature for object instance $i$ at time $t$, and let $\mathbf{g}_i^{\mathrm{text}}$ be the caption embedding. Caption alignment optimizes $\mathcal{L}_{\mathrm{cap}} = 1 - \cos(\phi(\mathbf{h}_{i,t}), \mathbf{g}_i^{\mathrm{text}})$, where $\phi$ is a projection head.

Fig.~\ref{fig:caption_alignment_diagnostic} diagnoses why this generic language-alignment strategy is insufficient for closed-set occupancy. The left panel evaluates the learned text-space representation: paired text cosine measures the optimized feature-caption similarity, nearest class-prototype accuracy tests whether the projected feature selects the correct closed-set class text prototype, and instance retrieval@5 tests whether it retrieves the paired crop caption. These metrics improve during training, showing that the auxiliary branch is learned. However, the right panel shows that caption alignment and closed-set semantic prototype distillation both remain below the reproduced occupancy baseline in semantic mIoU and occupied IoU. Thus, the issue is not simply an unoptimized auxiliary branch or the use of free-form captions; text-space supervision is weakly coupled to voxel-level closed-set semantic decisions. VISA therefore replaces caption embeddings with structured semantic audits that can be grounded to object voxels and distilled directly into semantic logits.

\textbf{Instance Semantic Audit Tuple:}
VISA uses the VLM as a structured semantic auditor rather than a caption generator. As mentioned in Section~\ref{sec:problem}, we parsed the output of the VLM auditor as an audit tuple $\mathcal{A}_i$, as defined in Eq.~\ref{eq:audit_tuple}.
The tuple is not used to supervise arbitrary scene regions; it is later grounded only to the corresponding 3D object voxels. 
The visual-factor vocabulary is predefined from occupancy-relevant driving semantics rather than freely generated by the VLM; examples include \texttt{large\_vehicle}, \texttt{long\_body}, \texttt{two\_wheeler}, \texttt{vulnerable\_road\_user}, and \texttt{traffic\_control\_object}. The full vocabulary is provided in Appendix~\ref{app:impl}. For example, a long vehicle crop may produce $c_i^{\mathrm{vlm}}=\texttt{bus}$, $\mathcal{C}_i=\{\texttt{truck},\texttt{trailer}\}$, and positive factors such as \texttt{large\_vehicle} and \texttt{long\_body}. Thus, each audit is taxonomy-aligned, instance-conditioned, reliability-aware, and directly usable for voxel-level training.

\textbf{Track-to-Voxel Grounding:}
A VLM audit is a 2D object-level observation, while occupancy supervision is applied to 3D voxel logits. VISA bridges this gap by associating each audit with the same physical object track across time and applying it only to the matched object-voxel set $\mathcal{V}_{i,t}$ defined in Eq.~\ref{eq:object_voxel_set}. This keeps crop-level VLM supervision local to the corresponding object and prevents leakage to background, free space, or nearby instances.

For each supervised object box, we pool the model predictions inside $\mathcal{V}_{i,t}$:
\begin{equation}
    \bar{\mathbf{p}}_{i,t}
    =
    \frac{1}{|\mathcal{V}_{i,t}|}
    \sum_{v \in \mathcal{V}_{i,t}}
    \mathrm{softmax}(\mathbf{z}_{v,t}).
\end{equation}
The dense occupancy loss still supervises the full voxel grid, while VISA losses act only on matched object regions. For example, a reliable audit from one visible crop of a bus is associated with the same bus track across time, but the loss is applied only to occupied voxels inside each corresponding 3D box.

\textbf{Reliability-Weighted Taxonomy Distillation:}
The taxonomy loss keeps the occupancy label as the anchor while allowing the VLM audit to express instance-specific ambiguity. Let $q_i$ be the VLM-reported confidence for its closed-set class hypothesis and $r_i$ be the crop reliability. We do not treat these scores as calibrated probabilities; they are used only for conservative filtering and weak loss weighting. An audit is used for taxonomy distillation only if $q_i \ge \tau_q$ and $r_i \ge \tau_r$.

For a valid audit, we construct a normalized VLM distribution $\mathbf{s}_i$ from the class hypothesis and its confusion set:
\[
    s_{i,c}
    \propto
    q_i \mathbb{1}[c = c_i^{\mathrm{vlm}}]
    +
    \sum_{(c', \gamma) \in \mathcal{C}_i}
    \gamma \mathbb{1}[c = c'] ,
\]
where $\gamma$ is the VLM-reported confidence for a plausible confusion class. We blend this distribution with the occupancy object label:
\[
    \mathbf{t}_i^{\mathrm{tax}}
    =
    (1-\alpha)\,\mathrm{onehot}(y_i^{\mathrm{obj}})
    +
    \alpha\,\mathbf{s}_i .
\]
For each audited object box, we apply a weighted soft cross-entropy:
\[
    \mathcal{L}_{\mathrm{tax}}
    =
    \frac{1}{|\mathcal{I}_{\mathrm{tax}}|}
    \sum_{(i,t)\in\mathcal{I}_{\mathrm{tax}}}
    w_i
    \left(
    -
    \sum_c
    t_{i,c}^{\mathrm{tax}}
    \log \bar{p}_{i,t,c}
    \right),
    \qquad
    w_i = r_i \max(q_i, 0.25),
\]
where $\mathcal{I}_{\mathrm{tax}}$ is the set of audited boxes that pass the confidence and reliability thresholds. This target expresses plausible confusions, such as bus--truck--trailer ambiguity, without replacing the occupancy label.

\textbf{Attribute-Factor Distillation:}
Taxonomy targets capture class-level ambiguity, but they do not explicitly supervise the visual factors shared by confusable classes. VISA therefore also distills the audited visual-factor vector $\mathbf{u}_i$. We define a fixed class-to-attribute matrix $\mathbf{M}\in\{0,1\}^{C_{\mathrm{obj}}\times A}$, where each row describes the attributes implied by an object class; for example, large vehicles share factors such as \texttt{large\_vehicle} and \texttt{long\_body}, while bicycles and motorcycles share \texttt{two\_wheeler}. The full mapping is provided in Appendix~\ref{app:impl}.
Let $\bar{\mathbf{p}}_{i,t}^{\mathrm{obj}}$ be the pooled object-class probability distribution over $\mathcal{V}_{i,t}$. We project it through $\mathbf{M}$ to obtain the predicted attribute vector $\hat{\mathbf{u}}_{i,t}=(\bar{\mathbf{p}}_{i,t}^{\mathrm{obj}})^\top\mathbf{M}$ and train it to match the VLM attribute vector:
\begin{equation}
    \mathcal{L}_{\mathrm{attr}}
    =
    \frac{1}{|\mathcal{I}_{\mathrm{attr}}|}
    \sum_{(i,t)\in\mathcal{I}_{\mathrm{attr}}}
    r_i\,
    \mathrm{BCE}(\hat{\mathbf{u}}_{i,t}, \mathbf{u}_i),
\end{equation}
where $\mathcal{I}_{\mathrm{attr}}$ contains reliable audited boxes with valid attribute targets. This loss encourages the semantic logits to respect instance-specific visual factors rather than only flat closed-set labels.

\textbf{Scene-Level Audit Graph:}
Taxonomy and attribute losses supervise audited instances independently. To add scene-level structure, VISA builds an audit graph over reliable audited objects at each timestep. Nodes are audited object instances with valid matched voxels and non-empty attribute vectors. Edges are assigned deterministically from symbolic audit fields rather than from VLM-provided edge scores.

For two nodes $i$ and $j$, let $\mathrm{sim}_{a}(\mathbf{u}_i,\mathbf{u}_j)$ be the Jaccard similarity of their binary attribute vectors, and let $\mathrm{compat}(i,j)$ indicate that the two audits have the same VLM class hypothesis or that one class appears in the other's confusion set. We define positive and negative edges as
\begin{align}
    (i,j) \in \mathcal{E}_t^{+}
    &\quad \mathrm{if} \quad
    \mathrm{sim}_{a}(\mathbf{u}_i,\mathbf{u}_j) \ge \tau_{+}
    \;\; \mathrm{or} \;\;
    \mathrm{compat}(i,j), \\
    (i,j) \in \mathcal{E}_t^{-}
    &\quad \mathrm{if} \quad
    \mathrm{sim}_{a}(\mathbf{u}_i,\mathbf{u}_j) \le \tau_{-}
    \;\; \mathrm{and} \;\;
    \neg \mathrm{compat}(i,j),
\end{align}
and ignore ambiguous pairs.

The model-side relation is computed from predicted attribute factors. Let
$\rho_{ij}=\cos(\hat{\mathbf{u}}_{i,t},\hat{\mathbf{u}}_{j,t})$ and
$w_{ij}=\min(r_i,r_j)$. The graph loss is
\begin{equation}
    \mathcal{L}_{\mathrm{graph}}
    =
    \frac{1}{Z}
    \left[
    \sum_{(i,j)\in\mathcal{E}_t^{+}}
    w_{ij}(1-\rho_{ij})
    +
    \sum_{(i,j)\in\mathcal{E}_t^{-}}
    w_{ij}\max(0,\rho_{ij}-m)
    \right],
\end{equation}
where $Z$ normalizes by the total valid edge weight. This encourages objects with compatible audits, such as buses and trucks, to share predicted semantic factors while separating incompatible pairs such as large vehicles and traffic cones.

\textbf{Training Objective and Inference:}
The VISA supervision term in Eq.~\ref{eq:objective_loss} combines local audit distillation and scene-level relation regularization:
\begin{equation}
    \mathcal{L}_{\mathrm{VISA}}
    =
    \lambda_{\mathrm{tax}}
    \mathcal{L}_{\mathrm{tax}}
    +
    \lambda_{\mathrm{attr}}
    \mathcal{L}_{\mathrm{attr}}
    +
    \lambda_{\mathrm{graph}}
    \mathcal{L}_{\mathrm{graph}} .
\end{equation}
All VLM audits are computed offline from training-set crops. At inference, the model receives only the standard camera inputs of the underlying occupancy world model and requires no image cropping, VLM querying, text encoder, object-track input, or additional module.
\section{Experiments}

\textbf{Experimental Setup:}
We evaluate VISA on nuScenes semantic occupancy~\cite{caesar2020nuscenes} with the standard 16 foreground classes. Following prior work~\cite{wei2023surroundocc,tian2023occ3d,huang2024gaussianformer,zuo2025gaussianworld}, we report occupied IoU, semantic mIoU, object mIoU over the 10 object classes, and rare mIoU over bicycle, bus, construction vehicle, motorcycle, traffic cone, and trailer; all numbers are percentages. We instantiate VISA on GaussianWorld and OccWorld under the same inputs, dense occupancy labels, training split, and evaluation protocol as each reproduced baseline, and report mean and standard deviation over three runs. The VLM is used only to generate offline training audits, and inference uses the original occupancy backbone. Additional implementation details are provided in Appendix~\ref{app:impl}.

\textbf{Main Benchmark Results:}
Table~\ref{tab:nuscenes_benchmark} compares VISA with prior camera-based semantic occupancy methods on nuScenes validation. Without changing the inference architecture, VISA improves semantic mIoU on both occupancy world-model backbones: OccWorld improves from $19.06$ to $20.05$, and GaussianWorld improves from $21.36$ to $21.91$, averaged across three runs. The gains are concentrated on object semantics, consistent with VISA's object-crop audit design. On GaussianWorld, object mIoU improves by $+0.98$ and rare-class mIoU improves by $+1.19$, with gains on visually confusable or long-tail classes such as bicycle, bus, construction vehicle, motorcycle, trailer, and truck. Stuff classes change less because VISA does not directly audit background regions.

\begin{table*}[t]
\centering
\caption{
NuScenes semantic occupancy benchmark.
IoU denotes binary occupied-space IoU and mIoU denotes semantic mIoU over 16 foreground classes.
All numbers are percentages.
For reproduced world-model backbones, IoU and mIoU are reported as mean $\pm$ standard deviation over three seeds, and per-class IoUs are averaged over the same seeds.
}
\label{tab:nuscenes_benchmark}
\setlength{\tabcolsep}{3.0pt}
\renewcommand{\arraystretch}{1.05}
\footnotesize
\begin{adjustbox}{max width=\textwidth}
\begin{tabular}{l cc cccccccccccccccc}
\toprule
\textbf{Method}
& \textbf{IoU}
& \textbf{mIoU}
& \occclass{orange}{barrier}
& \occclass{pink}{bicycle}
& \occclass{yellow}{bus}
& \occclass{cyan}{car}
& \occclass{teal}{const. veh.}
& \occclass{olive}{motorcycle}
& \occclass{red}{pedestrian}
& \occclass{orange!35}{traffic cone}
& \occclass{brown}{trailer}
& \occclass{violet}{truck}
& \occclass{magenta}{drive. surf.}
& \occclass{gray}{other flat}
& \occclass{purple!60!black}{sidewalk}
& \occclass{green!55}{terrain}
& \occclass{gray!35}{manmade}
& \occclass{green!70!black}{vegetation} \\
\midrule
MonoScene~\cite{cao2022monoscene}
& 23.96 & 7.31
& 4.03 & 0.35 & 8.00 & 8.04 & 2.90 & 0.28 & 1.16 & 0.67
& 4.01 & 4.35 & 27.72 & 5.20 & 15.13 & 11.29 & 9.03 & 14.86 \\
Atlas~\cite{murez2020atlas}
& 28.66 & 15.00
& 10.64 & 5.68 & 19.66 & 24.94 & 8.90 & 8.84 & 6.47 & 3.28
& 10.42 & 16.21 & 34.86 & 15.46 & 21.89 & 20.95 & 11.21 & 20.54 \\
BEVFormer~\cite{li2022bevformer}
& 30.50 & 16.75
& 14.22 & 6.58 & 23.46 & 28.28 & 8.66 & 10.77 & 6.64 & 4.05
& 11.20 & 17.78 & 37.28 & 18.00 & 22.88 & 22.17 & 13.80 & 22.21 \\
TPVFormer~\cite{huang2023tpvformer}
& 30.86 & 17.10
& 15.96 & 5.31 & 23.86 & 27.32 & 9.79 & 8.74 & 7.09 & 5.20
& 10.97 & 19.22 & 38.87 & 21.25 & 24.26 & 23.15 & 11.73 & 20.81 \\
OccFormer~\cite{zhang2023occformer}
& 31.39 & 19.03
& 18.65 & 10.41 & 23.92 & 30.29 & 10.31 & 14.19 & 13.59 & 10.13
& 12.49 & 20.77 & 38.78 & 19.79 & 24.19 & 22.21 & 13.48 & 21.35 \\
SurroundOcc~\cite{wei2023surroundocc}
& 31.49 & 20.30
& 20.59 & 11.68 & 28.06 & 30.86 & 10.70 & 15.14 & 14.09 & 12.06
& 14.38 & 22.26 & 37.29 & 23.70 & 24.49 & 22.77 & 14.89 & 21.86 \\
GaussianFormer~\cite{huang2024gaussianformer}
& 29.83 & 19.10
& 19.52 & 11.26 & 26.11 & 29.78 & 10.47 & 13.83 & 12.58 & 8.67
& 12.74 & 21.57 & 39.63 & 23.28 & 24.46 & 22.99 & 9.59 & 19.12 \\
GaussianFormer-2~\cite{huang2025gaussianformer2}
& 31.74 & 20.82
& 21.39 & 13.44 & 28.49 & 30.82 & 10.92 & 15.84 & 13.55 & 10.53
& 14.04 & 22.92 & 40.61 & 24.36 & 26.08 & 24.27 & 13.83 & 21.98 \\
QuadricFormer~\cite{zuo2026quadricformer}
& 32.13 & 21.11
& 21.38 & 13.41 & 28.40 & 31.01 & 11.32 & 17.10 & 13.94 & 11.28
& 14.75 & 22.66 & 40.81 & 24.71 & 26.51 & 25.22 & 13.54 & 21.78 \\
\midrule
OccWorld~\cite{zheng2023occworld}
& 31.63 $\pm$ 1.07 & 19.06 $\pm$ 0.29
& 15.55 & 13.38 & 25.16 & 27.49 & 10.08 & 13.67 & 12.44 & 9.52
& 9.16 & 17.83 & \textbf{42.34} & 24.58 & 25.32 & 23.95 & 10.97 & \textbf{23.48} \\
\textbf{OccWorld + VISA}
& \textbf{31.80} $\pm$ 1.94 & \textbf{20.05} $\pm$ 0.06
& \textbf{17.48} & \textbf{15.02} & \textbf{27.84} & \textbf{27.85} & \textbf{10.23} & \textbf{15.95} & \textbf{12.95} & \textbf{10.32}
& \textbf{10.10} & \textbf{20.58} & 41.52 & \textbf{25.37} & \textbf{25.83} & \textbf{24.20} & \textbf{12.59} & 23.01 \\
\midrule
GaussianWorld~\cite{zuo2025gaussianworld}
& 33.17 $\pm$ 1.13 & 21.36 $\pm$ 0.03
& 20.44 & 13.09 & 26.86 & 31.01 & 12.76 & 17.10 & 13.33 & \textbf{10.69}
& 14.11 & 22.94 & 42.31 & \textbf{25.26} & 27.53 & 25.81 & 14.38 & 24.21 \\
\textbf{GaussianWorld + VISA}
& \textbf{33.53} $\pm$ 1.02 & \textbf{21.91} $\pm$ 0.12
& \textbf{20.91} & \textbf{14.86} & \textbf{27.87} & \textbf{31.67} & \textbf{13.82} & \textbf{18.18} & \textbf{13.87} & 10.46
& \textbf{15.16} & \textbf{23.50} & \textbf{42.95} & 23.93 & \textbf{27.95} & \textbf{26.26} & \textbf{14.63} & \textbf{24.60} \\
\bottomrule
\end{tabular}
\end{adjustbox}
\vspace{-1.5em}
\end{table*}

\textbf{Qualitative World-State Comparison:}
\label{sec:qualitative}
\begin{figure*}[t]
    \centering
    \includegraphics[width=\textwidth]{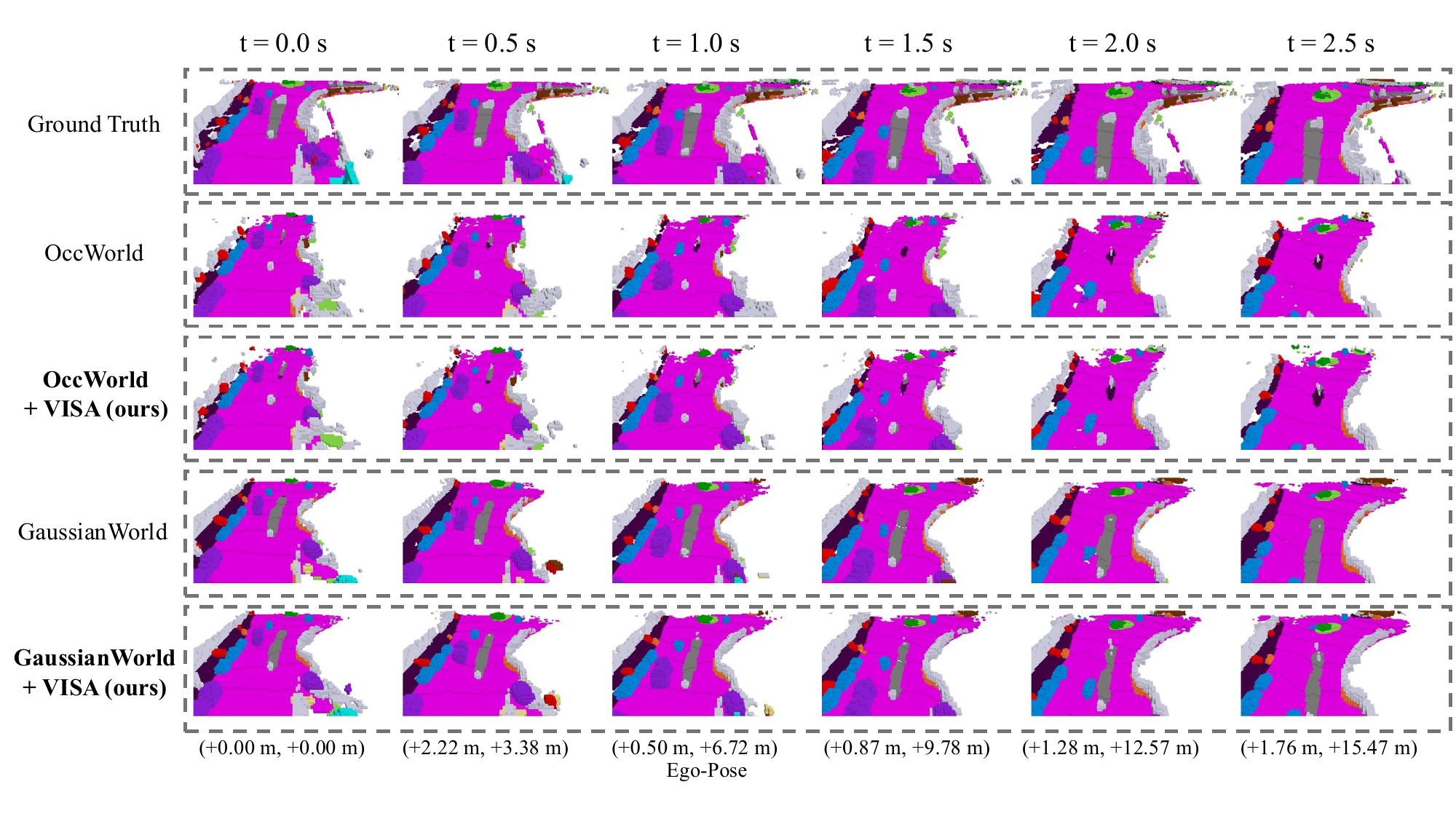}
      \vspace{-3em}
\caption{
Qualitative effect of VISA on occupancy world models.
We visualize the evolution of predicted occupancy world states from two backbones before and after adding VISA across six consecutive frames.
VISA yields cleaner foreground object regions and more coherent semantic boundaries while leaving the inference architecture unchanged.
}
    \label{fig:qual_visa_effect}
    \vspace{-1em}
\end{figure*}
Figure~\ref{fig:qual_visa_effect} compares the semantic occupancy states produced by baseline and VISA-enhanced models on the same nuScenes sequence. VISA does not change the inference architecture, but its training-time audits yield cleaner object regions and more coherent semantic boundaries across consecutive frames. These improvements are most visible around foreground objects under ego motion, where semantic drift or fragmentation can affect the quality of the maintained world state. This qualitative comparison is consistent with the object and rare-class gains in Table~\ref{tab:nuscenes_benchmark}.

\begin{figure*}[t]
\centering
\includegraphics[width=\textwidth]{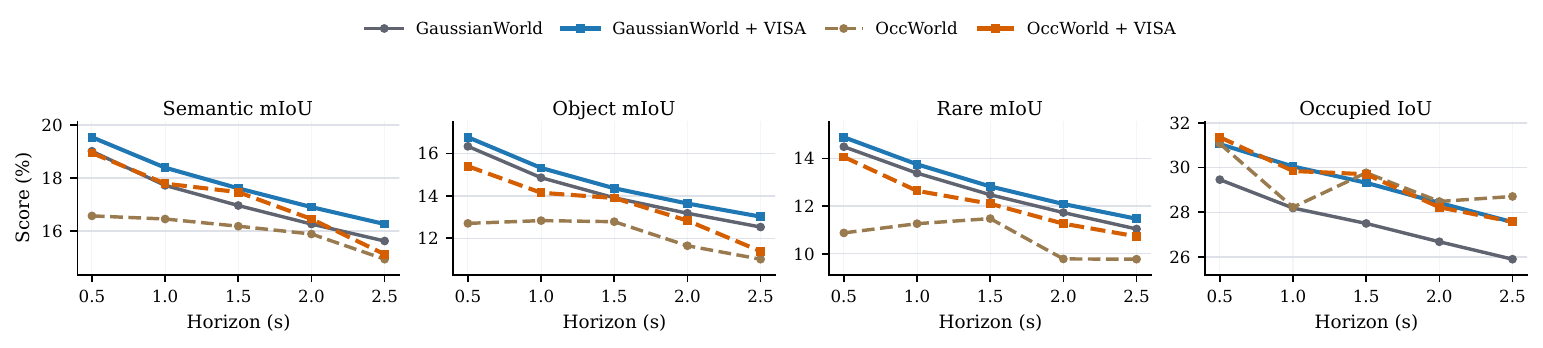}
 \vspace{-2.5em}
\caption{
Ego-motion-conditioned world-state rollout on nuScenes validation. Predicted occupancy states are propagated to future ego frames using known future ego poses. VISA consistently improves rollout quality on both GaussianWorld and OccWorld without using future images.
}
\label{fig:rollout}
 \vspace{-1.5em}
\end{figure*}

\begin{table*}[t]
\centering
\caption{
Ablation study on GaussianWorld. Left: component ablation of VISA. Right: non-VLM prior baselines that test whether the gains can be explained by taxonomy or attribute smoothing without instance-specific VLM auditing.
}
\label{tab:ablation_controls}
\begin{minipage}[t]{0.43\textwidth}
\centering
\centering
\setlength{\tabcolsep}{3.0pt}
\footnotesize
\resizebox{\linewidth}{!}{%
\begin{tabular}{ccc|cccc}
\toprule
\textbf{Tax.} & \textbf{Attr.} & \textbf{Graph}
& \textbf{IoU} & \textbf{mIoU} & \textbf{Obj.} & \textbf{Rare} \\
\midrule
\xmark & \xmark & \xmark & 33.29 & 21.37 & 18.18 & 15.60 \\
\cmark & \xmark & \xmark & \textbf{34.47} & 21.47 & 18.06 & 15.92 \\
\xmark & \cmark & \xmark & 32.00 & 21.40 & 18.85 & 16.47 \\
\cmark & \cmark & \xmark & 33.26 & 21.74 & 18.53 & 16.08 \\
\cmark & \cmark & \cmark & 33.61 & \textbf{21.91} & \textbf{19.16} & \textbf{16.79} \\
\bottomrule
\end{tabular}%
}

\vspace{0.35em}
\small (a) VISA components
\end{minipage}
\hspace{0.015\textwidth}
\begin{minipage}[t]{0.54\textwidth}
\centering
\centering
\setlength{\tabcolsep}{3.0pt}
\footnotesize
\resizebox{\linewidth}{!}{%
\begin{tabular}{l|cccc}
\toprule
\textbf{Prior baseline} & \textbf{IoU} & \textbf{mIoU} & \textbf{Obj.} & \textbf{Rare} \\
\midrule
GaussianWorld & 33.29 & 21.37 & 18.18 & 15.60 \\
+ Hand-crafted class neighbors & \textbf{34.44} & 21.38 & 17.34 & 15.06 \\
+ GT class attributes & 33.64 & 21.16 & 17.64 & 15.88 \\
+ Dataset confusion prior & 34.14 & 21.45 & 17.62 & 15.16 \\
+ Random confusion prior & 31.78 & 19.64 & 16.48 & 14.35 \\
\midrule
\textbf{+ VISA} & 33.61 & \textbf{21.91} & \textbf{19.16} & \textbf{16.79} \\
\bottomrule
\end{tabular}%
}

\vspace{0.35em}
\small (b) Non-VLM prior baselines
\vspace{0.35em}
\end{minipage}
\vspace{-2.5em}
\end{table*}

\textbf{Ego-Motion-Conditioned World-State Rollout:}
\label{sec:world_state_rollout}
Beyond current-frame occupancy, we evaluate whether the semantic world state predicted at time $t$ remains useful after propagation to future ego frames. Following the rollout setting of occupancy world models, we transform the predicted semantic occupancy state using known future ego poses and compare the propagated state with future occupancy labels. Figure~\ref{fig:rollout} shows that VISA improves rollout quality on both GaussianWorld and OccWorld. The gains are consistent across semantic mIoU, object mIoU, rare-class mIoU, and occupied IoU, indicating that instance-level VLM auditing improves the semantic quality of the world state before rollout. The improvement is most pronounced for object and rare-class metrics, matching the object-centric design of VISA.

\textbf{Ablations and Non-VLM Controls:}
\label{sec:ablation}
Table~\ref{tab:ablation_controls} evaluates VISA components and non-VLM alternatives on GaussianWorld. In the component ablation, taxonomy supervision improves semantic mIoU and occupied IoU, while attribute-factor supervision improves object and rare-class mIoU. Combining taxonomy and attributes improves overall mIoU but does not fully preserve the object/rare gains of attribute-only supervision. Adding the scene-level audit graph gives the best semantic, object, and rare-class results, suggesting that relational audit regularization helps stabilize the combination of taxonomy and attribute cues. The non-VLM controls test whether VISA reduces to class-structured label smoothing. Hand-crafted taxonomy smoothing, GT-derived attributes, and dataset-confusion smoothing use the same backbone and training schedule but remove instance-specific VLM audits; random-confusion smoothing controls for the soft-label format. These alternatives either improve occupied IoU while hurting object/rare mIoU, or degrade all metrics. VISA is the only variant that improves semantic, object, and rare-class mIoU together, indicating that the gains come from instance-conditioned visual audits rather than global taxonomy priors alone.
\begin{wraptable}{r}{0.48\linewidth}
\vspace{-0.5em}
\centering
\caption{
Offline VLM audit diagnostics on the nuScenes training split.
}
\label{tab:audit_diagnostics}
\setlength{\tabcolsep}{4pt}
\footnotesize
\begin{tabular}{l|c}
\toprule
\textbf{Statistic} & \textbf{Value} \\
\midrule
Training-scene coverage & 700 / 700 \\
Audited 3D-box instances & 51,420 \\
Evaluated object-class audits & 49,568 \\
Closed-set agreement & 87.7\% \\
Clear / partial / poor crops & 20.7 / 76.6 / 2.7\% \\
Rare-class audit fraction & 15.1\% \\
\bottomrule
\end{tabular}
\vspace{-2em}
\end{wraptable}

\textbf{Offline VLM Audit Diagnostics:}
Table~\ref{tab:audit_diagnostics} summarizes the offline audits used by VISA. The audit pipeline covers all 700 nuScenes training scenes and produces 51,420 audited 3D-box instances, including 49,568 audits from the 10 object classes used for object mIoU. The VLM closed-set class hypothesis agrees with the nuScenes label for 87.7\% of evaluated object-class audits, showing that the teacher is generally aligned with the occupancy taxonomy but is not accurate enough to be treated as an oracle.
The visibility breakdown further supports reliability-weighted supervision. Clear, partial, and poor crops account for 20.7\%, 76.6\%, and 2.7\% of sampled audits, with top-1 agreement of 92.9\%, 88.6\%, and 22.9\%, respectively. Their average confidence/reliability scores are 0.95/0.96, 0.66/0.60, and 0.18/0.14. VISA therefore uses VLM outputs as soft, reliability-weighted taxonomy and attribute supervision, while dense occupancy labels remain the primary supervision over the full voxel grid.

\label{sec:audit_diagnostics}

\section{Conclusion, Limitations and Future Work}
\label{sec:conclusion}


\textbf{Conclusion:}
We presented \textbf{VISA}, a training-time semantic auditing framework for existing 3D occupancy world models. VISA converts offline VLM judgments on object crops into closed-set, voxel-grounded supervision through taxonomy, attribute-factor, and scene-level graph losses, while leaving inference VLM-free. On nuScenes, VISA improves OccWorld and GaussianWorld, especially on object and rare-class semantics. Our blind audit diagnostics further show that the VLM provides meaningful visual evidence without access to ground-truth class labels. These results suggest that VLMs are more useful for closed-set occupancy as reliability-aware semantic auditors than as generic caption-embedding targets.

\textbf{Limitations and Future Work:}
First, VISA adds offline audit-generation cost and can inherit recognition biases from the VLM teacher. This cost is paid only during training, but scaling to larger datasets requires efficient audit generation and teacher-quality validation. Future work can reduce this dependency through lighter VLMs, active audit selection, or cached reusable audit banks. Second, VISA currently audits visible object instances, while stuff regions such as road, sidewalk, terrain, and vegetation remain supervised mainly by dense occupancy labels. This matches the object-centric gains observed in our experiments, but limits the benefit for large background regions and scene layout. Extending semantic auditing from object crops to stuff regions or scene-level context is an important direction. Third, VISA relies on annotated object boxes/tracks and dense voxel labels during training to ground crop-level audits to matched object voxels. These annotations are not used at inference, but they limit the direct applicability of the current implementation to datasets with reliable temporal object annotations. Using detected tracks or weaker instance annotations is an important direction for scaling VISA.

\clearpage
\appendix
\section{Additional Related Work}
\label{app:related}
\textbf{3D Semantic Occupancy and Occupancy World Models:}
Originating from semantic scene completion~\cite{song2017sscnet,cao2022monoscene}, 3D semantic occupancy prediction has become a standard representation for surround-view robot perception~\cite{wang2023openoccupancy,tian2023occ3d}.
Mainstream architectures construct volumetric scene fields by lifting multi-camera features into voxel or BEV-voxel representations with spatiotemporal transformers, view transformation modules, or efficient 3D decoders~\cite{li2022bevformer,zhang2023occformer,li2023fbocc,wang2024panoocc}.
Parallel efforts explore scalable supervision beyond dense 3D labels, including temporal consistency across frames~\cite{wei2023surroundocc,huang2024selfocc} and differentiable rendering from 2D observations~\cite{pan2024renderocc,zhang2023occnerf}.
More recently, occupancy representations have been incorporated into world-model formulations, where the predicted 3D state can be propagated or rolled out for future scene reasoning and planning~\cite{ha2018worldmodels,zheng2023occworld,zuo2025gaussianworld}.
However, closed-set semantic occupancy remains vulnerable to long-tail imbalance and fine-grained category confusion, especially for object classes whose 2D evidence is sparse, occluded, or visually ambiguous.
VISA complements temporal aggregation and world-state rollout by improving the semantic quality of the learned occupancy state, without changing the underlying world-model architecture or inference pipeline.

\textbf{Vision-Language Supervision for 3D Robot Perception:}
Transferring knowledge from 2D vision-language models (VLMs) to 3D perception has progressed from sparse point clouds~\cite{peng2023openscene,yang2023regionplc} to dense voxel prediction.
Recent language-driven occupancy methods typically follow two paradigms.
One line aligns 3D features with continuous language or CLIP embeddings, sometimes through differentiable rendering or voxel-language contrastive learning~\cite{tan2023ovo,vobecky2024pop3d,boeder2024langocc,zhang2025clip}.
Another line uses foundation models to generate dense pseudo-labels or open-vocabulary supervision offline, followed by 2D--3D fusion or occupancy distillation~\cite{yu2024language,zheng2024veon,doruk2026vlmfusion,feng2024vipocc}.
While effective for open-vocabulary perception, these strategies are not always aligned with closed-set occupancy evaluation: embedding alignment may require high-dimensional language-aligned features and can suffer from cross-modal representation mismatch, while dense pseudo-label generation can introduce boundary noise and projection errors in voxel space.
In contrast, VISA uses the VLM as an offline semantic auditor rather than an inference-time feature extractor or dense pseudo-labeler.
It converts structured crop-level audits into closed-set taxonomy, attribute, and scene-graph supervision over voxel logits, introducing no VLM calls or language features at inference.

\textbf{Taxonomy- and Attribute-Aware Robot Supervision:}
Real-world driving categories exhibit semantic relationships and shared physical properties, but standard occupancy models usually treat classes as flat and independent labels.
Hierarchical semantic parsing in 2D vision~\cite{li2022deep} suggests that taxonomic structure can improve semantic consistency, yet directly building hierarchical architectures for dense 3D voxel prediction can increase model complexity.
Existing 3D occupancy methods more commonly address imbalance and ambiguity through class-balanced objectives~\cite{cui2019classbalanced,kang2019decoupling}, contrastive learning~\cite{khosla2020supervised}, auxiliary geometric guidance~\cite{Zhu2026DrOccDA}, or active sample selection~\cite{Kim2026ClassDistributionGA,Leng2025OccupancyLW}.
These approaches do not explicitly model visual factors shared across related objects, such as long-body vehicles, two-wheelers, vulnerable road users, or traffic-control obstacles.
As a result, visually similar but functionally distinct categories can remain difficult to separate, particularly for rare or safety-critical classes~\cite{tian2023occ3d,Mattamala2024WildVN}.
VISA addresses this gap by using VLM audits to provide taxonomy-aware class relations and attribute-factor supervision during training, improving rare and visually ambiguous object semantics while preserving the original inference-time world model.



\section{Implementation Details}
\label{app:impl}

\paragraph{Crop extraction and view selection.}
We generate VLM audits only from the nuScenes training split.
For each annotated 3D object box, we project the box into all six surround-view cameras and compute a tight 2D bounding box whenever at least two projected corners are visible.
To reduce severe occlusion, we compute inter-box overlap in each camera view and mark a projected crop as occluded when its visible area ratio is below $0.3$.
We select the best camera view by preferring unoccluded projections with larger visible crop area; if all views are occluded, we fall back to the best available projection.
The final crop is padded by $10\%$ of the tight 2D box size with a minimum padding of 8 pixels, clipped to the image boundary, and saved as a JPEG.
We discard crops whose tight 2D box is smaller than 24 pixels on either side.
Each crop record stores the scene, frame index, object box index, selected camera, tight and padded 2D boxes, depth, visibility ratio, occlusion flag, number of LiDAR points, and instance token when available.

\paragraph{Blind VLM audit prompt.}
The VLM is not given the ground-truth class label.
Instead, we query Qwen3-VL offline with only the crop image, a closed-set occupancy object vocabulary, and a fixed visual-factor vocabulary.
The prompt is designed to elicit structured semantic audits rather than free-form captions.
The core prompt is:

\begin{quote}
\small
You are auditing a small crop from a vehicle surround-view camera.
Choose only from this closed-set class list:
barrier, bicycle, bus, car, construction\_vehicle, motorcycle, pedestrian, traffic\_cone, trailer, truck.
Your task is not to write a caption.
Output an instance-level taxonomy and visual-factor audit using only visible crop evidence:
the most likely closed-set class, confidence, up to three plausible confusion classes, a visual subtype, binary visual attributes useful for 3D semantic occupancy, visual reliability, visibility, motion likelihood, and one short visual reason.
Do not use scene context.
For visibility, use clear when the class is identifiable from the crop, partial when the crop is truncated, occluded, tiny, or blurred but still contains class evidence, and poor when the crop is too ambiguous to identify from visual evidence.
Output only one JSON object with the required fields.
\end{quote}

The required JSON fields are:
\texttt{nuScenes\_class}, \texttt{class\_confidence}, \texttt{possible\_confusions}, \texttt{subtype}, \texttt{attributes}, \texttt{motion\_likelihood}, \texttt{visibility}, \texttt{visual\_reliability}, and \texttt{reason}.
The attribute keys are fixed to:
vehicle-like, large vehicle, long body, box-shaped object, two-wheeler, vulnerable road user, traffic-control object, construction-like object, static obstacle, and thin vertical object.
Each attribute is binary and must be supported by visible evidence.

\paragraph{JSON validation and audit statistics.}
We parse only valid JSON objects and canonicalize class names to the 10 object classes used for object mIoU.
Class confidence and reliability are clipped to $[0,1]$; malformed confusion entries, invalid fields, or missing attributes are replaced by conservative defaults.
The full blind audit covers all 700 nuScenes training scenes and produces 51,420 audited 3D-box instances, including 49,568 instances from the 10 evaluated object classes.
The parse-valid rate after JSON extraction and validation is $100.0\%$.
The merged audit file is generated once offline and reused for all VISA training runs.

\paragraph{Blind audit quality diagnostic.}
Because the VLM never receives the ground-truth class label, we can directly evaluate its closed-set visual agreement with nuScenes annotations.
Overall, the blind audit achieves $87.7\%$ top-1 agreement with nuScenes labels.
Agreement is strongly correlated with the VLM reliability estimate: clear crops have $92.9\%$ agreement, partial crops have $88.6\%$ agreement, and poor crops drop to $22.9\%$ agreement.
The corresponding mean confidence/reliability values are $0.95/0.96$ for clear crops, $0.66/0.60$ for partial crops, and $0.18/0.14$ for poor crops.
This supports using the VLM as a reliability-weighted semantic auditor rather than an oracle pseudo-labeler.

\paragraph{Voxel grounding and filtering.}
During training, each audit is attached to the matching physical object using the nuScenes instance token when available; otherwise it is matched by scene, frame index, and box index.
VISA applies audit supervision only to voxels inside the corresponding oriented 3D box whose dense occupancy label matches the object class.
We enlarge the box by a factor of $1.05$ to tolerate discretization error.
For taxonomy supervision, we keep audits with visual reliability at least $0.20$ and class confidence at least $0.20$.
For attribute and scene-graph supervision, we keep audits with visual reliability at least $0.20$.
We require at least 4 valid object voxels per box, supervise at most 80 boxes per frame for taxonomy and attribute losses, at most 60 boxes per frame for the scene graph loss, sample at most 4096 voxels per box for instance losses, and at most 2048 voxels per box for graph loss.

\paragraph{Training losses.}
VISA is added to the original occupancy objective as auxiliary training supervision.
The taxonomy loss constructs a soft target
\[
\tilde{\mathbf{y}}_i = (1-\alpha)\mathbf{e}_{y_i} + \alpha \mathbf{q}_i,
\]
where $\mathbf{e}_{y_i}$ is the closed-set occupancy label, $\mathbf{q}_i$ is the normalized VLM class/confusion distribution, and $\alpha=0.20$.
The taxonomy loss weight is $0.03$.
The attribute loss maps predicted object-class probabilities to the fixed attribute vocabulary and applies binary cross-entropy to the VLM attribute vector with weight $0.02$.
The scene-level audit graph builds pairwise relations between audited objects in the same frame: pairs are positive if their attribute Jaccard similarity is at least $0.50$ or their VLM taxonomy/confusion sets are compatible, and negative if their attribute Jaccard similarity is at most $0.05$ and the taxonomy sets are incompatible.
The negative similarity margin is $0.20$, the graph loss weight is $0.005$, and at most 512 object pairs are sampled per frame.
All VISA losses are applied only during training.

\paragraph{Offline cost and inference.}
The full audit generation is embarrassingly parallel across crops.
In our implementation, Qwen3-VL-8B is run with greedy decoding and up to 340 generated tokens per crop.
The audit bank covers 51,420 training 3D-box instances and is generated once offline before occupancy training.
At deployment, VISA introduces no additional inputs, parameters, VLM calls, text encoders, crop extraction, object tracks, or latency; inference is exactly the original occupancy world model.

\section{More Ablation Studies}

\begin{table}[t]
\centering
\caption{
Robustness to object grounding quality on GaussianWorld.
All variants use the same offline VLM audits and training schedule.
}
\label{tab:grounding_robustness}
\begin{tabular}{lcccc}
\toprule
Grounding setting & IoU & mIoU & Obj. & Rare \\
\midrule
Full VISA: GT boxes/tracks & 33.61 & 21.91 & 19.16 & 16.79 \\
w/o track broadcast & 32.81 & 21.34 & 18.42 & 15.92 \\
Jittered boxes & 33.25 & 21.67 & 18.81 & 16.34 \\
w/o voxel-label filtering & 32.07 & 21.21 & 17.98 & 15.47 \\
\bottomrule
\end{tabular}
\end{table}

\paragraph{Robustness to object grounding quality.}
Table~\ref{tab:grounding_robustness} studies how VISA depends on the quality of object-level grounding.
The full model uses GT 3D boxes, instance tracks, and voxel-label filtering to attach each blind VLM audit to matched object voxels.
Removing track broadcast reduces mIoU from 21.91 to 21.34 and rare mIoU from 16.79 to 15.92, showing that propagating an audit along the same physical object track improves supervision coverage, especially for long-tail classes.
When the 3D boxes used for VISA grounding are jittered with moderate noise, performance remains close to the full model, with mIoU decreasing by only 0.24 and rare mIoU by 0.45.
This suggests that VISA is not overly sensitive to small box localization errors.
Finally, removing voxel-label filtering gives the largest degradation in object and rare-class metrics.
This confirms that restricting VLM supervision to voxels whose dense label matches the audited object class prevents semantic leakage from nearby stuff regions or neighboring objects inside the 3D box.
Overall, the results show that VISA benefits from accurate object grounding, but remains robust to moderate box noise, while track broadcast and matched-voxel filtering are important for clean and effective supervision.

\begin{table}[t]
\centering
\caption{
Ablation of audit reliability and crop selection on GaussianWorld.
All variants use the same training schedule and offline VLM audit format.
}
\label{tab:audit_reliability_crop}
\setlength{\tabcolsep}{4.5pt}
\renewcommand{\arraystretch}{1.06}
\footnotesize
\begin{tabular}{lcccc}
\toprule
\textbf{Variant} & \textbf{IoU} & \textbf{mIoU} & \textbf{Obj.} & \textbf{Rare} \\
\midrule
VISA full
& \textbf{33.61} & \textbf{21.91} & \textbf{19.16} & \textbf{16.79} \\
w/o reliability weighting
& 33.35 & 21.63 & 18.84 & 16.35 \\
w/o reliability filtering
& 31.90 & 20.81 & 17.10 & 15.51 \\
Random visible crop
& 32.08 & 21.32 & 18.23 & 15.68 \\
Largest-area crop
& 33.48 & 21.74 & 17.98 & 16.53 \\
\bottomrule
\end{tabular}
\end{table}

\paragraph{Audit reliability.}
Table~\ref{tab:audit_reliability_crop} first evaluates the reliability mechanism in VISA.
Removing reliability weighting consistently reduces performance, lowering mIoU from 21.91 to 21.63 and rare mIoU from 16.79 to 16.35.
This indicates that the VLM's confidence and visual reliability scores provide useful information for modulating the strength of audit supervision.
Removing reliability filtering causes a larger degradation, especially on object mIoU, which drops from 19.16 to 17.10.
This suggests that low-quality or ambiguous audits can introduce harmful semantic noise if they are allowed to supervise voxel logits.
Thus, reliability is not merely a diagnostic output of the VLM; it is an important part of making VLM supervision robust.

\paragraph{Crop selection.}
Table~\ref{tab:audit_reliability_crop} also studies how the choice of audited crop affects training.
Using a random visible crop degrades IoU and rare mIoU, showing that arbitrary views can produce noisier audits and weaker supervision.
The largest-area crop baseline recovers much of the overall mIoU and rare-class performance, but still underperforms full VISA in object mIoU.
This suggests that visible area is an important factor for audit quality, but area alone is not sufficient: the full crop-selection strategy also accounts for occlusion and representative visibility, which improves object-level semantic supervision.
\clearpage
\begin{center}
{\LARGE \textbf{Supplementary Material}}\\[0.5em]
{\large \textbf{VISA: VLM-Guided Instance Semantic Auditing for 3D Occupancy World Models}}
\end{center}
\vspace{1em}
\appendix
\section{Future Occupancy Forecasting Diagnostics}
\label{sec:supp_future_forecasting}

We further evaluate how the semantic occupancy representations learned with VISA affect future occupancy forecasting.
This experiment is intended as an additional temporal diagnostic beyond single-frame occupancy evaluation.
Given a stream occupancy backbone, we train the same lightweight forecasting-head architecture to predict future occupancy states from past predicted occupancy states.
The forecasting head is trained on predicted states rather than ground-truth occupancy, so the evaluation reflects the compounded errors of the perception backbone and the forecasting module.
At test time, we autoregressively roll out predictions from 0.5s to 2.5s and report semantic mIoU, object mIoU, rare-class mIoU, and occupied-space IoU on the nuScenes validation set.
All methods use the same forecasting-head training protocol, validation split, horizons, and metric definitions.
All metrics are reported in percentages.

\begin{figure*}[t]
\centering
\includegraphics[width=0.88\textwidth]{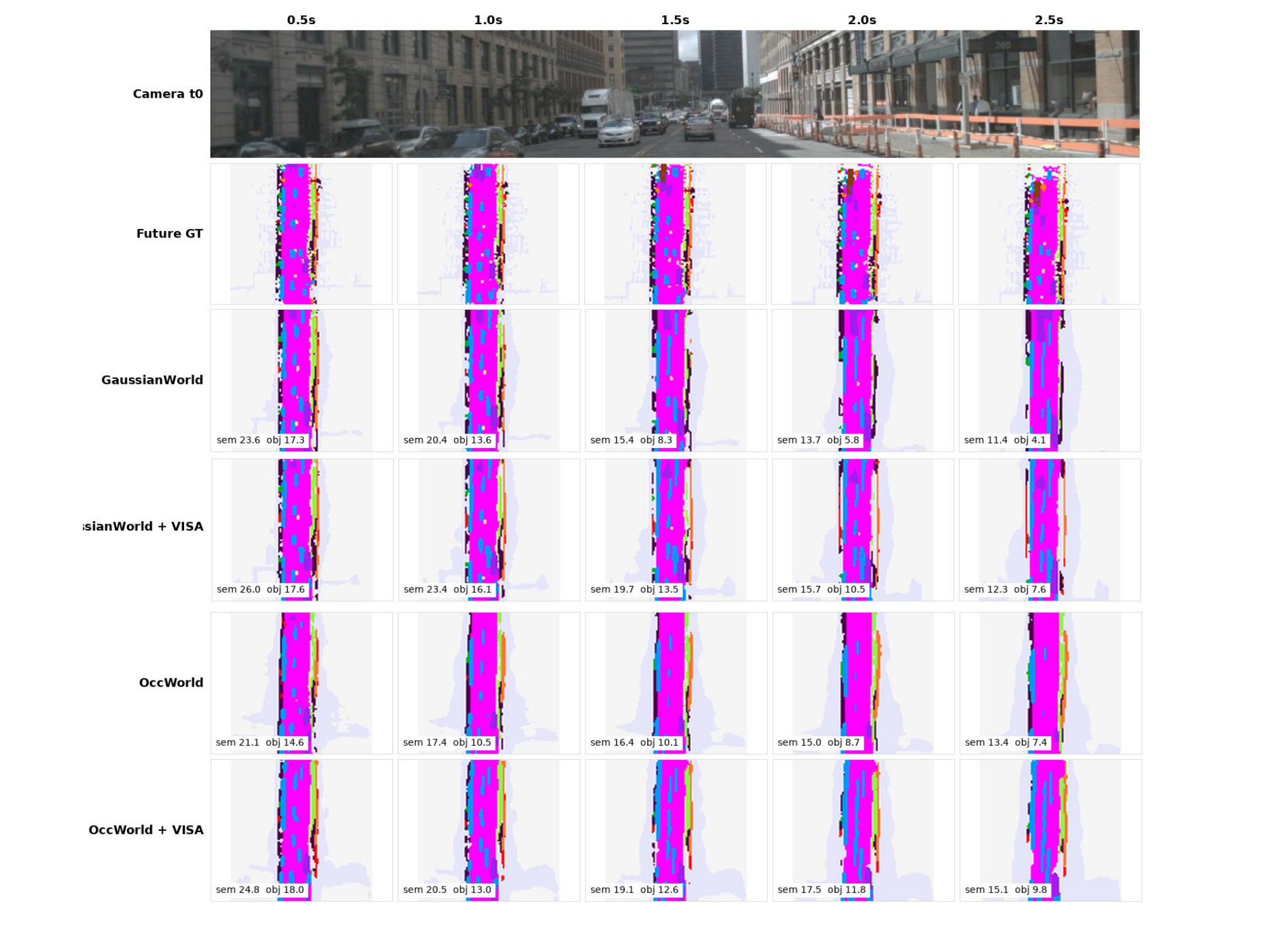}
\caption{
Qualitative future occupancy forecasting visualization on nuScenes validation.
The columns show forecasting horizons from 0.5s to 2.5s, and rows compare future ground truth with predictions from GaussianWorld, GaussianWorld+VISA, OccWorld, and OccWorld+VISA.
The numbers under each prediction denote semantic and object scores for the displayed forecast.
In this representative sequence, VISA preserves sharper semantic boundaries and cleaner object-centric foreground regions over longer horizons.
}
\label{fig:future_forecasting_qualitative}
\end{figure*}

\begin{table*}[t]
\centering
\caption{
Future occupancy forecasting diagnostics from 0.5s to 2.5s on nuScenes validation.
Sem., Obj., Rare, and Occ. IoU denote semantic mIoU, object mIoU, rare-class mIoU,
and occupied-space IoU, respectively. All metrics are reported in percentages;
values in parentheses indicate percentage-point changes relative to the corresponding reproduced backbone.
Bold numbers indicate the better result within each backbone pair at the same horizon.
}
\label{tab:future_forecasting_all}
\scriptsize
\setlength{\tabcolsep}{3pt}
\resizebox{\textwidth}{!}{
\begin{tabular}{llcccc}
\toprule
Horizon & Method & Sem. (\%) $\uparrow$ & Obj. (\%) $\uparrow$ & Rare (\%) $\uparrow$ & Occ. IoU (\%) $\uparrow$ \\
\midrule
\multirow{4}{*}{0.5s}
& GaussianWorld & 21.03 & \textbf{15.60} & \textbf{10.12} & 37.89 \\
& GaussianWorld + VISA & \textbf{21.07} {\scriptsize(+0.04)} & 15.48 {\scriptsize(-0.12)} & 9.89 {\scriptsize(-0.23)} & \textbf{38.56} {\scriptsize(+0.67)} \\
& OccWorld & 15.74 & 11.29 & 6.28 & 32.37 \\
& OccWorld + VISA & \textbf{16.87} {\scriptsize(+1.13)} & \textbf{12.57} {\scriptsize(+1.28)} & \textbf{7.28} {\scriptsize(+1.00)} & \textbf{32.67} {\scriptsize(+0.30)} \\
\midrule
\multirow{4}{*}{1.0s}
& GaussianWorld & 18.54 & \textbf{11.94} & \textbf{7.01} & 36.17 \\
& GaussianWorld + VISA & \textbf{18.68} {\scriptsize(+0.14)} & 11.51 {\scriptsize(-0.43)} & 6.89 {\scriptsize(-0.12)} & \textbf{37.42} {\scriptsize(+1.25)} \\
& OccWorld & 14.17 & 8.57 & 4.30 & \textbf{32.16} \\
& OccWorld + VISA & \textbf{14.97} {\scriptsize(+0.80)} & \textbf{9.54} {\scriptsize(+0.97)} & \textbf{4.80} {\scriptsize(+0.50)} & 32.11 {\scriptsize(-0.05)} \\
\midrule
\multirow{4}{*}{1.5s}
& GaussianWorld & 16.27 & \textbf{9.37} & 5.02 & 33.54 \\
& GaussianWorld + VISA & \textbf{16.75} {\scriptsize(+0.48)} & 9.04 {\scriptsize(-0.33)} & \textbf{5.32} {\scriptsize(+0.30)} & \textbf{35.75} {\scriptsize(+2.21)} \\
& OccWorld & 12.78 & 6.78 & 3.11 & \textbf{31.42} \\
& OccWorld + VISA & \textbf{13.19} {\scriptsize(+0.41)} & \textbf{7.58} {\scriptsize(+0.80)} & \textbf{3.43} {\scriptsize(+0.32)} & 30.35 {\scriptsize(-1.07)} \\
\midrule
\multirow{4}{*}{2.0s}
& GaussianWorld & 14.22 & 7.34 & 3.58 & 30.89 \\
& GaussianWorld + VISA & \textbf{15.17} {\scriptsize(+0.95)} & \textbf{7.41} {\scriptsize(+0.07)} & \textbf{4.24} {\scriptsize(+0.66)} & \textbf{33.93} {\scriptsize(+3.04)} \\
& OccWorld & 11.52 & 5.43 & 2.11 & \textbf{30.41} \\
& OccWorld + VISA & \textbf{11.67} {\scriptsize(+0.15)} & \textbf{6.28} {\scriptsize(+0.85)} & \textbf{2.63} {\scriptsize(+0.52)} & 28.23 {\scriptsize(-2.18)} \\
\midrule
\multirow{4}{*}{2.5s}
& GaussianWorld & 12.46 & 5.89 & 2.80 & 28.51 \\
& GaussianWorld + VISA & \textbf{13.77} {\scriptsize(+1.31)} & \textbf{6.11} {\scriptsize(+0.22)} & \textbf{3.39} {\scriptsize(+0.59)} & \textbf{32.14} {\scriptsize(+3.63)} \\
& OccWorld & \textbf{10.34} & 4.36 & 1.37 & \textbf{29.24} \\
& OccWorld + VISA & 10.31 {\scriptsize(-0.03)} & \textbf{5.20} {\scriptsize(+0.84)} & \textbf{1.92} {\scriptsize(+0.55)} & 26.15 {\scriptsize(-3.09)} \\
\bottomrule
\end{tabular}
}
\end{table*}

Figure~\ref{fig:future_forecasting_qualitative} provides a qualitative example of the same forecasting setting.
In this challenging urban driving scene, VISA produces visibly sharper future occupancy forecasts, with cleaner foreground regions and more coherent semantic boundaries over time.
The improvement is consistent across the displayed horizons: both GaussianWorld+VISA and OccWorld+VISA achieve higher semantic and object scores than their corresponding backbones from 0.5s to 2.5s.
This visualization highlights the benefit of injecting instance-level semantic structure into the occupancy representation before autoregressive future prediction.

Table~\ref{tab:future_forecasting_all} shows that the effect of VISA on future occupancy forecasting is metric- and backbone-dependent.
For GaussianWorld, VISA improves semantic mIoU and occupied-space IoU across all horizons.
The gains become larger at longer horizons: at 2.5s, GaussianWorld+VISA improves semantic mIoU by 1.31 points and occupied-space IoU by 3.63 points.
Object and rare-class forecasting are slightly lower at short horizons, but improve at longer horizons.
This suggests that, for GaussianWorld, the occupancy states learned with VISA provide a more favorable input representation for long-horizon forecasting.

For OccWorld, VISA shows a different pattern.
It improves object mIoU at every horizon, with gains between 0.80 and 1.28 points, and also improves rare-class mIoU across all horizons.
This is consistent with the object-centric design of VISA and indicates that the learned representation helps preserve object and rare-class semantic information during future prediction.
However, occupied-space IoU decreases at longer horizons for OccWorld+VISA.
This indicates that improvements in object-level semantic discrimination do not necessarily translate into improved coarse occupied support under autoregressive forecasting.

Overall, these results show that VISA can provide benefits beyond static occupancy evaluation, but the effect is not uniform across backbones or metrics.
On GaussianWorld, the benefits are strongest for long-horizon semantic and geometric forecasting.
On OccWorld, the benefits are concentrated on object and rare-class forecasting, while coarse geometric occupancy becomes worse at longer horizons.
This behavior is consistent with VISA's design: it injects instance- and semantic-level structure into occupancy representations, which is most directly useful for preserving object-class information over time, but it does not directly optimize future free-space geometry or coarse occupied support.

\section{Open-Loop Planning and Near-Field Semantic Diagnostics}
\label{sec:supp_planning_diagnostics}

We further evaluate whether the semantic occupancy representations learned with VISA benefit downstream open-loop planning diagnostics and near-field semantic perception.
This experiment is intended as a diagnostic rather than a closed-loop safety evaluation.
For open-loop planning, we use the same downstream planning protocol for each reproduced backbone and train the same lightweight planning-head architecture.
All methods follow the same planning-head training and evaluation protocol.
We report trajectory L2 error in meters at the 1s, 2s, and 3s horizons, as well as the average L2 across horizons.
For transparency, we also report the mean instantaneous box-collision rate over the same horizons.

To connect the planning diagnostic with semantic occupancy quality, we report near-field object mIoU and rare-class mIoU within 20m of the ego vehicle.
These near-field metrics are computed directly from the semantic occupancy predictions and are reported in percentages.
All near-field evaluations use the same validation scenes, frame count, and metric definitions.

\begin{table*}[t]
\centering
\caption{
Open-loop planning and near-field semantic occupancy diagnostics on nuScenes validation.
L2 errors are reported in meters at the 1s, 2s, and 3s horizons, and Avg. L2 denotes their average.
BoxCol is the mean instantaneous box-collision rate over the same horizons.
Near-field metrics are computed within 20m of the ego vehicle and reported in percentages.
Values in parentheses indicate changes relative to the corresponding backbone; for L2 columns, changes are in meters, and for percentage metrics, changes are in percentage points.
Bold numbers indicate the better result within each backbone pair.
}
\label{tab:planning_l2_nearfield}
\scriptsize
\setlength{\tabcolsep}{3pt}
\resizebox{\textwidth}{!}{
\begin{tabular}{lcccc|c|cc}
\toprule
Method
& L2@1s $\downarrow$
& L2@2s $\downarrow$
& L2@3s $\downarrow$
& Avg. L2 $\downarrow$
& BoxCol (\%) $\downarrow$
& Obj.@20 (\%) $\uparrow$
& Rare@20 (\%) $\uparrow$ \\
\midrule
GaussianWorld
& 0.956
& 1.479
& 2.064
& 1.500
& \textbf{4.79}
& 28.40
& 24.93 \\
GaussianWorld + VISA
& \textbf{0.859} {\scriptsize(-0.097)}
& \textbf{1.406} {\scriptsize(-0.073)}
& \textbf{2.013} {\scriptsize(-0.051)}
& \textbf{1.426} {\scriptsize(-0.074)}
& 4.99 {\scriptsize(+0.20)}
& \textbf{28.41} {\scriptsize(+0.01)}
& \textbf{25.19} {\scriptsize(+0.26)} \\
\midrule
OccWorld
& 0.882
& 1.462
& 2.072
& 1.472
& \textbf{4.28}
& 21.12
& 17.69 \\
OccWorld + VISA
& \textbf{0.842} {\scriptsize(-0.040)}
& \textbf{1.425} {\scriptsize(-0.037)}
& \textbf{2.048} {\scriptsize(-0.024)}
& \textbf{1.438} {\scriptsize(-0.034)}
& 4.97 {\scriptsize(+0.69)}
& \textbf{22.78} {\scriptsize(+1.66)}
& \textbf{18.89} {\scriptsize(+1.20)} \\
\bottomrule
\end{tabular}
}
\end{table*}

Table~\ref{tab:planning_l2_nearfield} shows that VISA reduces open-loop trajectory L2 for both reproduced occupancy world-model backbones under the same downstream planning protocol.
For GaussianWorld, VISA reduces average L2 from 1.500 to 1.426.
The improvement is consistent across all horizons, with reductions of 0.097m at 1s, 0.073m at 2s, and 0.051m at 3s.
For OccWorld, VISA reduces average L2 from 1.472 to 1.438, with smaller but consistent gains across the three horizons.
These results suggest that the semantic occupancy states learned with VISA can improve expert-trajectory imitation in an open-loop planning setting.

The near-field semantic diagnostics show a consistent object-centric trend.
VISA improves object mIoU within 20m for both backbones, with a small gain on GaussianWorld and a larger gain on OccWorld.
It also improves rare-class mIoU within 20m for both backbones.
These gains are aligned with the design of VISA, which provides training-time supervision on matched object voxels and is intended to improve object and rare-class semantic quality in planning-relevant regions.

The box-collision diagnostic shows a different trend: mean BoxCol does not improve under this open-loop planning protocol.
This reflects the difference between expert-trajectory imitation and collision-aware planning.
L2 measures similarity to logged trajectories, whereas box collision depends on footprint clearance, geometric support, and planner safety costs.
We therefore interpret this section as an open-loop planning and near-field semantic diagnostic rather than a closed-loop safety evaluation.
VISA does not directly optimize planner collision cost, clearance margins, or closed-loop control behavior.
The results indicate that VISA improves open-loop trajectory imitation and near-field object-centric semantic occupancy, but should not be interpreted as evidence of improved collision avoidance or closed-loop planning safety.

\section{Additional VLM Audit Diagnostics}
\label{sec:supp_vlm_audit_diagnostics}

VISA uses an offline VLM auditor to provide training-time instance-level semantic structure.
Since this supervision is derived from a vision-language model, we further analyze the quality and ambiguity patterns of the audit signal used during training.
For each object crop, the auditor returns a structured response containing a closed-set class hypothesis, plausible confusion classes, visual attributes, class confidence, visual reliability, and short evidence.
We report audit agreement with the nuScenes class annotation, as well as reliability-binned audit quality.
These diagnostics are intended to characterize the training audit bank used by VISA, rather than to benchmark the VLM as a standalone classifier.

\begin{table}[t]
\centering
\caption{
VLM audit quality on the rare classes used for rare-class mIoU.
Top-1 agreement measures whether the VLM's predicted class matches the nuScenes annotation.
In-conf. agreement measures whether the ground-truth class is either the top prediction or appears in the VLM-proposed confusion set.
Conf. and Rel. denote the average VLM-reported class confidence and visual reliability.
}
\label{tab:vlm_audit_per_class}
\setlength{\tabcolsep}{5pt}
\begin{tabular}{lccccc}
\toprule
Class & \# Audits & Top-1 (\%) & In-conf. (\%) & Conf. & Rel. \\
\midrule
Bicycle & 574 & 67.3 & 76.9 & 0.66 & 0.63 \\
Bus & 523 & 76.7 & 93.3 & 0.76 & 0.72 \\
Constr. vehicle & 529 & 60.0 & 95.0 & 0.86 & 0.79 \\
Motorcycle & 605 & 86.3 & 88.2 & 0.62 & 0.61 \\
Traffic cone & 4617 & 91.0 & 97.1 & 0.78 & 0.73 \\
Trailer & 917 & 48.3 & 79.3 & 0.84 & 0.83 \\
\bottomrule
\end{tabular}
\end{table}

Table~\ref{tab:vlm_audit_per_class} shows that the audit signal is informative but not noise-free.
For visually distinctive categories such as traffic cone and motorcycle, the VLM often predicts the correct closed-set class.
For more ambiguous rare classes such as trailer and construction vehicle, top-1 agreement is lower, but the ground-truth class frequently appears in the proposed confusion set.
This is important for VISA because the audit is not used as a hard replacement label.
Instead, VISA uses the full structured audit, including plausible confusions, visual attributes, confidence, and reliability, as soft object-centric supervision anchored by the original occupancy labels.
The confidence and reliability values should therefore be interpreted as audit metadata for weighting and filtering, not as guarantees that the top-1 VLM class is semantically correct.

\begin{table}[t]
\centering
\caption{
Audit quality as a function of VLM-reported visual reliability over the full training audit set.
Reliability is not treated as a calibrated probability, but higher reliability bins correlate with higher audit agreement.
}
\label{tab:vlm_audit_reliability_bins}
\setlength{\tabcolsep}{6pt}
\begin{tabular}{lccc}
\toprule
Reliability bin & \# Audits & Top-1 (\%) & Avg. confidence \\
\midrule
$[0.0, 0.2)$ & 801 & 29.2 & 0.10 \\
$[0.2, 0.4)$ & 3153 & 79.9 & 0.29 \\
$[0.4, 0.6)$ & 8989 & 82.9 & 0.30 \\
$[0.6, 0.8)$ & 21198 & 90.4 & 0.76 \\
$[0.8, 1.0]$ & 17279 & 91.1 & 0.94 \\
\bottomrule
\end{tabular}
\end{table}

Table~\ref{tab:vlm_audit_reliability_bins} supports the use of reliability filtering and weighting in VISA.
Very low-reliability crops have poor agreement, while crops with reliability above 0.6 achieve over 90\% top-1 agreement.
Reliability is therefore useful as a training-time visual-quality signal, even though it should not be interpreted as a calibrated probability or as a guarantee of semantic correctness.
This distinction matters because a crop can be visually clear and thus receive high reliability, while still being semantically ambiguous under the fine-grained closed-set driving taxonomy.
In the full training audit bank, we filter or downweight low-quality signals using both class-confidence and visual-reliability thresholds.

\begin{table}[t]
\centering
\caption{
Representative structured VLM audit examples from the training audit set.
Ambiguous cases often involve visually similar object categories, but the ground-truth class is frequently preserved in the VLM confusion set or reflected in the visual attributes.
}
\label{tab:vlm_audit_examples}
\setlength{\tabcolsep}{4pt}
\begin{tabular}{lllccl}
\toprule
Type & GT & VLM & Confid. & Rel. & Notes \\
\midrule
Correct & Constr. vehicle & Constr. vehicle & 0.98 & 0.99 & construction-like, large vehicle \\
Correct & Bicycle & Bicycle & 0.95 & 0.98 & two-wheeler, vulnerable road user \\
Correct & Traffic cone & Traffic cone & 0.95 & 0.98 & traffic-control, thin obstacle \\
\midrule
Ambiguous & Constr. vehicle & Truck & 0.98 & 0.98 & GT appears in confusion set \\
Ambiguous & Trailer & Truck & 0.95 & 0.98 & long-body vehicle ambiguity \\
Ambiguous & Traffic cone & Barrier & 0.95 & 0.98 & traffic-control obstacle ambiguity \\
\bottomrule
\end{tabular}
\end{table}

Table~\ref{tab:vlm_audit_examples} illustrates why VISA uses structured audits rather than single hard pseudo-labels.
The dominant audit errors are not random class assignments, but structured confusions among visually related driving categories, such as trailer versus truck, construction vehicle versus truck, and traffic cone versus barrier.
Some of these ambiguous cases can still have high visual reliability because the crop itself is clear, even though the fine-grained closed-set category remains ambiguous.
This motivates the VISA design: reliability is used to filter poor visual evidence, while the confusion set and attribute factors capture uncertainty among semantically related classes.
The audit signal is used only during training as auxiliary supervision; inference remains VLM-free.

\section{Offline Audit Generation Cost}
\label{sec:supp_offline_audit_cost}

VISA uses the VLM auditor only as an offline training-data preprocessing step.
The generated audit bank is fixed before occupancy training and is not queried during validation or inference.
As a result, deployment is unchanged relative to the corresponding occupancy backbone: no VLM module is attached, no image-text model is loaded, no crop extraction is performed, and no VLM call is made at test time.

\begin{table}[t]
\centering
\caption{
Offline VLM audit generation cost on the nuScenes training audit set.
Audit generation is parallel over object crops and is run once before occupancy training.
Throughput and time are measured from our sharded generation logs on NVIDIA RTX A6000 GPUs.
The audit bank is used only during training and introduces no additional inference-time modules or VLM calls.
}
\label{tab:offline_audit_cost}
\setlength{\tabcolsep}{4pt}
\resizebox{\linewidth}{!}{
\begin{tabular}{ll}
\toprule
Item & Value \\
\midrule
VLM backbone & Qwen3-VL-8B-Instruct \\
\# audited crops & 51,420 \\
GPU type & NVIDIA RTX A6000 \\
Generation setup & 4 nodes $\times$ 4 GPUs/node, 16 GPUs total \\
Decoding setting & Greedy decoding, max 340 new tokens, structured audit prompt \\
Throughput & 0.128 crops/s/GPU \\
Total wall-clock time & 7.1 hours \\
Total GPU-hours & 111.3 GPU-hours \\
Audit bank storage & 113 MB JSONL; 293 MB including manifests, shards, and logs \\
JSON schema-valid records & 100.0\% \; (51,420 / 51,420) \\
Additional inference modules & None \\
VLM calls at inference & None \\
Additional inference overhead & None beyond the original occupancy backbone \\
\bottomrule
\end{tabular}
}
\end{table}

Table~\ref{tab:offline_audit_cost} summarizes the one-time cost of generating the offline audit bank.
The cost is paid before occupancy training and is naturally parallelizable because each object crop can be audited independently.
In our implementation, each GPU processes a disjoint JSONL shard and writes structured audit records without distributed synchronization.
This makes the preprocessing simple to scale with the number of available GPUs.

The resulting audit bank is compact compared with model checkpoints and dataset storage.
For each crop, it stores the audited class hypothesis, plausible confusion classes, visual attributes, confidence, reliability, and the raw structured response.
During occupancy training, VISA reads these records to construct the taxonomy, attribute, and graph supervision losses.
At inference time, the audit records and VISA losses are removed entirely.
Thus, VISA preserves the deployed occupancy model architecture and input pipeline: it does not require an image-text model, VLM queries, object crops, object tracks, or additional inference-time memory and latency.

\section{Failure Cases and Boundary Conditions}
\label{sec:supp_failure_cases}

We provide qualitative failure cases and boundary conditions to clarify what VISA does and does not address.
These examples are intentionally selected stress cases rather than typical outcomes.
They are meant to illustrate the operating limits of VLM-guided semantic auditing and structured occupancy supervision.
VISA uses offline VLM audits as soft training-time semantic guidance, rather than treating the VLM as an oracle or adding inference-time reasoning modules.
The examples highlight two complementary sources of residual error: ambiguity in crop-level audit signals and local errors in spatially coherent semantic occupancy prediction.

\begin{figure}[t]
\centering
\includegraphics[width=\linewidth]{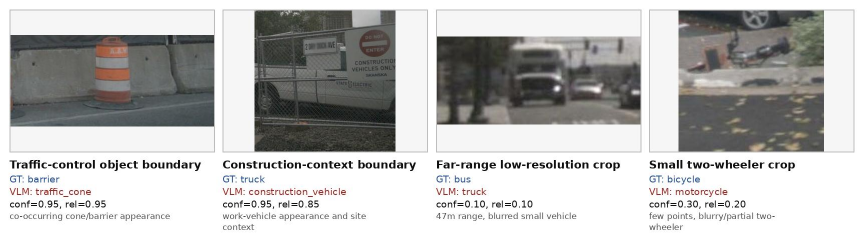}
\caption{
Representative VLM audit boundary cases from the full training audit set.
The examples illustrate foreground distractors, contextual bias, low-resolution evidence, and viewpoint ambiguity.
We report the VLM class confidence and visual reliability for each crop.
}
\label{fig:audit_boundary_cases}
\end{figure}

Figure~\ref{fig:audit_boundary_cases} shows representative failure modes of the offline VLM audit.
In the first example, the target instance is a concrete barrier, but a salient traffic cone appears in the foreground; the VLM focuses on the foreground object and predicts \emph{traffic cone} instead of \emph{barrier}.
In the second example, the target truck is partially visible behind a fence, and the surrounding sign and construction-like context bias the audit toward \emph{construction vehicle}.
The third example is a far-range, low-resolution crop, where the vehicle shape is too blurred to reliably distinguish \emph{bus} from \emph{truck}.
The fourth example contains a fallen two-wheeler viewed from above, where the visible evidence indicates a two-wheeler but is insufficient to reliably separate \emph{bicycle} from \emph{motorcycle}.

These cases show that best-view crop selection improves visibility but cannot completely remove foreground distractors, contextual bias, far-range resolution limits, or viewpoint ambiguity.
This is why VISA does not use VLM predictions as hard pseudo-labels.
Instead, each audit is used as a soft, reliability-aware training signal, anchored by the original dense occupancy supervision and restricted to matched object voxels.
Low-confidence or low-reliability audits are filtered or weakly weighted, while even reliable audits remain auxiliary guidance rather than replacements for ground-truth occupancy labels.

\begin{figure}[t]
\centering
\includegraphics[width=\linewidth]{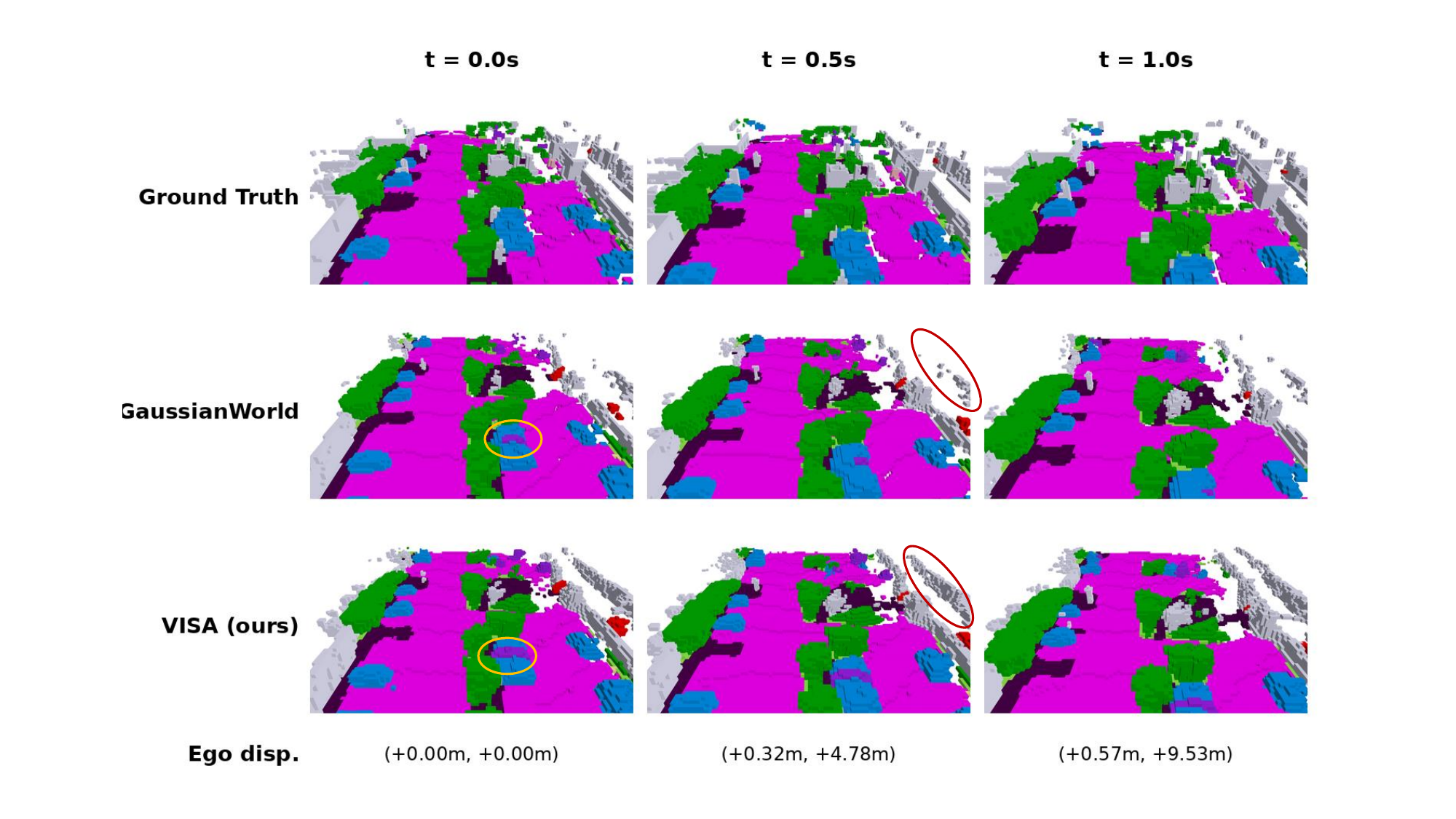}
\caption{
Representative boundary case of structured occupancy completion on nuScenes validation.
Red circles highlight over-completed regions where the prediction fills spatially coherent occupancy that is absent from the ground truth.
Yellow circles highlight cases where a local semantic misclassification is propagated to nearby voxels due to the same structured completion behavior.
Highlighted regions are representative rather than exhaustive.
}
\label{fig:occupancy_boundary_case}
\end{figure}

Figure~\ref{fig:occupancy_boundary_case} shows a boundary case of spatially coherent semantic occupancy prediction.
VISA often produces cleaner and more coherent object or region-level predictions, which is beneficial when the local semantic evidence is correct.
However, in locally ambiguous regions, this same coherence can make an incorrect local hypothesis appear as a more contiguous false-positive region, as highlighted by the red circles.
Similarly, when a small region receives an incorrect semantic label, the prediction can become spatially more coherent while remaining semantically incorrect over a local neighborhood, as highlighted by the yellow circles.

This behavior reflects a boundary of consistency-oriented semantic supervision rather than an additional inference-time failure mode.
VISA encourages object-centric semantic consistency, but it does not directly solve all sources of geometric uncertainty, stuff-region ambiguity, or local semantic noise.
Thus, when the underlying local evidence is incorrect or incomplete, spatial coherence can improve visual consistency without necessarily correcting the underlying semantic error.

Overall, these stress cases help explain the remaining errors observed despite the quantitative gains reported in the main paper.
They clarify the regime in which VISA is most useful: improving object-centric semantic occupancy supervision, especially for rare and visually confusable classes, while preserving the original inference architecture.
The method does not fully resolve ambiguous crop-level grounding, contextual VLM bias, low-resolution evidence, viewpoint ambiguity, or local errors caused by incomplete geometric evidence.
These limitations motivate future work on uncertainty-aware audit selection, stronger geometric consistency checks, and extending semantic auditing beyond visible object instances to stuff regions and free-space structure.


\clearpage


\bibliography{example}  

@inproceedings{song2017sscnet,
  title={Semantic Scene Completion from a Single Depth Image},
  author={Song, Shuran and Yu, Fisher and Zeng, Andy and Chang, Angel X. and Savva, Manolis and Funkhouser, Thomas},
  booktitle={CVPR},
  year={2017},
  url={https://arxiv.org/abs/1611.08974}
}

@inproceedings{caesar2020nuscenes,
  title={nuScenes: A Multimodal Dataset for Autonomous Driving},
  author={Caesar, Holger and Bankiti, Varun and Lang, Alex H. and Vora, Sourabh and Liong, Venice Erin and Xu, Qiang and Krishnan, Anush and Pan, Yu and Baldan, Giancarlo and Beijbom, Oscar},
  booktitle={CVPR},
  year={2020},
  url={https://arxiv.org/abs/1903.11027}
}

@inproceedings{cao2022monoscene,
  title={MonoScene: Monocular 3D Semantic Scene Completion},
  author={Cao, Anh-Quan and de Charette, Raoul},
  booktitle={CVPR},
  year={2022},
  url={https://arxiv.org/abs/2112.00726}
}

@inproceedings{li2022bevformer,
  title={BEVFormer: Learning Bird's-Eye-View Representation from Multi-Camera Images via Spatiotemporal Transformers},
  author={Li, Zhiqi and Wang, Wenhai and Li, Hongyang and Xie, Enze and Sima, Chonghao and Lu, Tong and Yu, Qiao and Dai, Jifeng},
  booktitle={ECCV},
  year={2022},
  url={https://arxiv.org/abs/2203.17270}
}

@inproceedings{li2023voxformer,
  title={VoxFormer: Sparse Voxel Transformer for Camera-based 3D Semantic Scene Completion},
  author={Li, Yiming and Yu, Zhiding and Choy, Christopher B. and Xiao, Chaowei and Alvarez, Jose M. and Fidler, Sanja and Feng, Chen and Anandkumar, Anima},
  booktitle={CVPR},
  year={2023},
  url={https://arxiv.org/abs/2302.12251}
}

@inproceedings{huang2023tpvformer,
  title={Tri-Perspective View for Vision-Based 3D Semantic Occupancy Prediction},
  author={Huang, Yuanhui and Zheng, Wenzhao and Zhang, Yunpeng and Zhou, Jie and Lu, Jiwen},
  booktitle={CVPR},
  year={2023},
  url={https://arxiv.org/abs/2302.07817}
}

@inproceedings{wei2023surroundocc,
  title={SurroundOcc: Multi-Camera 3D Occupancy Prediction for Autonomous Driving},
  author={Wei, Yi and Zhao, Linqing and Zheng, Wenzhao and Zhu, Zheng and Zhou, Jie and Lu, Jiwen},
  booktitle={ICCV},
  year={2023},
  url={https://arxiv.org/abs/2303.09551}
}

@inproceedings{tian2023occ3d,
  title={Occ3D: A Large-Scale 3D Occupancy Prediction Benchmark for Autonomous Driving},
  author={Tian, Xiaoyu and Jiang, Tao and Yun, Longfei and Mao, Yue and Yang, Huitong and Wang, Yue and Wang, Yilun and Zhao, Hang},
  booktitle={NeurIPS Datasets and Benchmarks},
  year={2023},
  url={https://arxiv.org/abs/2304.14365}
}

@article{ha2018worldmodels,
  title={World Models},
  author={Ha, David and Schmidhuber, J{\"u}rgen},
  journal={arXiv preprint arXiv:1803.10122},
  year={2018},
  url={https://arxiv.org/abs/1803.10122}
}

@article{zheng2023occworld,
  title={OccWorld: Learning a 3D Occupancy World Model for Autonomous Driving},
  author={Zheng, Wenzhao and Chen, Weiliang and Huang, Yuanhui and Zhang, Borui and Duan, Yueqi and Lu, Jiwen},
  journal={arXiv preprint arXiv:2311.16038},
  year={2023},
  url={https://arxiv.org/abs/2311.16038}
}

@article{min2024driveworld,
  title={DriveWorld: 4D Pre-trained Scene Understanding via World Models for Autonomous Driving},
  author={Min, Chen and Xiao, Liang and Nie, Yizhou and Dai, Bin and Zhang, Songhai and others},
  journal={arXiv preprint arXiv:2405.04390},
  year={2024},
  url={https://arxiv.org/abs/2405.04390}
}

@inproceedings{radford2021clip,
  title={Learning Transferable Visual Models From Natural Language Supervision},
  author={Radford, Alec and Kim, Jong Wook and Hallacy, Chris and Ramesh, Aditya and Goh, Gabriel and Agarwal, Sandhini and Sastry, Girish and Askell, Amanda and Mishkin, Pamela and Clark, Jack and Krueger, Gretchen and Sutskever, Ilya},
  booktitle={International Conference on Machine Learning},
  year={2021},
  url={https://arxiv.org/abs/2103.00020}
}

@inproceedings{zhai2023siglip,
  title={Sigmoid Loss for Language Image Pre-Training},
  author={Zhai, Xiaohua and Mustafa, Basil and Kolesnikov, Alexander and Beyer, Lucas},
  booktitle={Proceedings of the IEEE/CVF International Conference on Computer Vision},
  pages={11975--11986},
  year={2023},
  url={https://arxiv.org/abs/2303.15343}
}

@inproceedings{li2023blip2,
  title={{BLIP}-2: Bootstrapping Language-Image Pre-training with Frozen Image Encoders and Large Language Models},
  author={Li, Junnan and Li, Dongxu and Savarese, Silvio and Hoi, Steven},
  booktitle={International Conference on Machine Learning},
  pages={19730--19742},
  year={2023},
  url={https://proceedings.mlr.press/v202/li23q.html}
}

@inproceedings{liu2023llava,
  title={Visual Instruction Tuning},
  author={Liu, Haotian and Li, Chunyuan and Wu, Qingyang and Lee, Yong Jae},
  booktitle={Advances in Neural Information Processing Systems},
  year={2023},
  url={https://arxiv.org/abs/2304.08485}
}

@article{bai2023qwenvl,
  title={{Qwen-VL}: A Versatile Vision-Language Model for Understanding, Localization, Text Reading, and Beyond},
  author={Bai, Jinze and Bai, Shuai and Yang, Shusheng and Wang, Shijie and Tan, Sinan and Wang, Peng and Lin, Junyang and Zhou, Chang and Zhou, Jingren},
  journal={arXiv preprint arXiv:2308.12966},
  year={2023},
  url={https://arxiv.org/abs/2308.12966}
}

@article{wang2024qwen2vl,
  title={{Qwen2-VL}: Enhancing Vision-Language Model's Perception of the World at Any Resolution},
  author={Wang, Peng and Bai, Shuai and Tan, Sinan and Wang, Shijie and Fan, Zhihao and Bai, Jinze and Chen, Keqin and Liu, Xuejing and Wang, Jialin and Ge, Wenbin and Fan, Yang and Dang, Kai and Du, Mengfei and Ren, Xuancheng and Men, Rui and Liu, Dayiheng and Zhou, Chang and Zhou, Jingren and Lin, Junyang},
  journal={arXiv preprint arXiv:2409.12191},
  year={2024},
  url={https://arxiv.org/abs/2409.12191}
}

@inproceedings{chen2023clip2scene,
  title={{CLIP2Scene}: Towards Label-Efficient 3D Scene Understanding by {CLIP}},
  author={Chen, Runnan and Liu, Youquan and Kong, Lingdong and Zhu, Xinge and Ma, Yuexin and Li, Yikang and Hou, Yuenan and Qiao, Yu and Wang, Wenping},
  booktitle={Proceedings of the IEEE/CVF Conference on Computer Vision and Pattern Recognition},
  pages={7020--7030},
  year={2023},
  url={https://openaccess.thecvf.com/content/CVPR2023/html/Chen_CLIP2Scene_Towards_Label-Efficient_3D_Scene_Understanding_by_CLIP_CVPR_2023_paper.html}
}

@inproceedings{peng2023openscene,
  title={OpenScene: 3D Scene Understanding With Open Vocabularies},
  author={Peng, Songyou and Genova, Kyle and Jiang, Chiyu Max and Tagliasacchi, Andrea and Pollefeys, Marc and Funkhouser, Thomas},
  booktitle={Proceedings of the IEEE/CVF Conference on Computer Vision and Pattern Recognition},
  pages={815--824},
  year={2023},
  url={https://openaccess.thecvf.com/content/CVPR2023/html/Peng_OpenScene_3D_Scene_Understanding_With_Open_Vocabularies_CVPR_2023_paper.html}
}

@article{ding2022pla,
  title={{PLA}: Language-Driven Open-Vocabulary 3D Scene Understanding},
  author={Ding, Runyu and Yang, Jihan and Xue, Chuhui and Zhang, Wenqing and Bai, Song and Qi, Xiaojuan},
  journal={arXiv preprint arXiv:2211.16312},
  year={2022},
  url={https://arxiv.org/abs/2211.16312}
}

@article{yang2023regionplc,
  title={RegionPLC: Regional Point-Language Contrastive Learning for Open-World 3D Scene Understanding},
  author={Yang, Jihan and Ding, Runyu and Deng, Weipeng and Wang, Zhe and Qi, Xiaojuan},
  journal={arXiv preprint arXiv:2304.00962},
  year={2023},
  url={https://arxiv.org/abs/2304.00962}
}

@inproceedings{chen2024internvl,
  title={{InternVL}: Scaling up Vision Foundation Models and Aligning for Generic Visual-Linguistic Tasks},
  author={Chen, Zhe and Wu, Jiannan and Wang, Wenhai and Su, Weijie and Chen, Guo and Xing, Sen and Muyan, Zhongzhi and Zhang, Qinglong and Zhu, Xizhou and Lu, Lewei and Li, Bin and Luo, Ping and Lu, Tong and Qiao, Yu and Dai, Jifeng},
  booktitle={Proceedings of the IEEE/CVF Conference on Computer Vision and Pattern Recognition},
  year={2024},
  url={https://arxiv.org/abs/2312.14238}
}

@article{li2024llavanext,
  title={{LLaVA-NeXT}: Improved Reasoning, OCR, and World Knowledge},
  author={Li, Bo and Zhang, Yuanhan and Guo, Dong and Zhang, Renrui and Li, Feng and Zhang, Hao and Zhang, Kaichen and Li, Peiyuan and Liu, Yanwei and Li, Chunyuan},
  journal={arXiv preprint arXiv:2401.13601},
  year={2024},
}

@inproceedings{huang2024gaussianformer,
  title={GaussianFormer: Scene as Gaussians for Vision-Based 3D Semantic Occupancy Prediction},
  author={Huang, Yuanhui and Zheng, Wenzhao and Zhang, Yunpeng and Zhou, Jie and Lu, Jiwen},
  booktitle={European Conference on Computer Vision},
  year={2024},
  url={https://arxiv.org/abs/2405.17429}
}

@inproceedings{zuo2025gaussianworld,
  title={GaussianWorld: Gaussian World Model for Streaming 3D Occupancy Prediction},
  author={Zuo, Sicheng and Zheng, Wenzhao and Huang, Yuanhui and Zhou, Jie and Lu, Jiwen},
  booktitle={Proceedings of the IEEE/CVF Conference on Computer Vision and Pattern Recognition},
  year={2025},
  url={https://arxiv.org/abs/2412.10373}
}

@inproceedings{cui2019classbalanced,
  title={Class-Balanced Loss Based on Effective Number of Samples},
  author={Cui, Yin and Jia, Menglin and Lin, Tsung-Yi and Song, Yang and Belongie, Serge},
  booktitle={Proceedings of the IEEE/CVF Conference on Computer Vision and Pattern Recognition},
  pages={9268--9277},
  year={2019},
  url={https://arxiv.org/abs/1901.05555}
}

@inproceedings{kang2019decoupling,
  title={Decoupling Representation and Classifier for Long-Tailed Recognition},
  author={Kang, Bingyi and Xie, Saining and Rohrbach, Marcus and Yan, Zhicheng and Gordo, Albert and Feng, Jiashi and Kalantidis, Yannis},
  booktitle={International Conference on Learning Representations},
  year={2020},
  url={https://arxiv.org/abs/1910.09217}
}

@inproceedings{khosla2020supervised,
  title={Supervised Contrastive Learning},
  author={Khosla, Prannay and Teterwak, Piotr and Wang, Chen and Sarna, Aaron and Tian, Yonglong and Isola, Phillip and Maschinot, Aaron and Liu, Ce and Krishnan, Dilip},
  booktitle={Advances in Neural Information Processing Systems},
  year={2020},
  url={https://arxiv.org/abs/2004.11362}
}

@article{tan2023ovo,
  title={OVO: Open-Vocabulary Occupancy},
  author={Tan, Zhiyu and Dong, Zichao and Zhang, Cheng-Jun and Zhang, Weikun and Ji, Hang and Li, Hao},
  journal={arXiv preprint arXiv:2305.16133},
  year={2023},
  url={https://arxiv.org/abs/2305.16133}
}

@article{vobecky2024pop3d,
  title={POP-3D: Open-Vocabulary 3D Occupancy Prediction from Images},
  author={Vobeck{\'y}, Anton{\'i}n and Sim'eoni, Oriane and Hurych, David and Gidaris, Spyros and Bursuc, Andrei and P{\'e}rez, Patrick and Sivic, Josef},
  journal={arXiv preprint arXiv:2401.09413},
  year={2024},
  url={https://arxiv.org/abs/2401.09413}
}

@article{boeder2024langocc,
  title={LangOcc: Self-Supervised Open Vocabulary Occupancy Estimation via Volume Rendering},
  author={Boeder, Simon and Gigengack, Fabian and Risse, Benjamin},
  journal={arXiv preprint arXiv:2407.17310},
  year={2024},
  url={https://arxiv.org/abs/2407.17310}
}

@inproceedings{yu2024language,
  title={Language Driven Occupancy Prediction},
  author={Yu, Zhu and Pang, Bowen and Liu, Lizhe and Zhang, Runmin and Peng, Qihao and Luo, Maochun and Yang, Sheng and Chen, Mingxia and Cao, Sixi and Shen, Hui},
  booktitle={Proceedings of the IEEE/CVF International Conference on Computer Vision (ICCV)},
  year={2024},
  url={https://arxiv.org/abs/2411.16072}
}

@article{zheng2024veon,
  title={VEON: Vocabulary-Enhanced Occupancy Prediction},
  author={Zheng, Jilai and Tang, Pin and Wang, Zhongdao and Wang, Guoqing and Ren, Xiangxuan and Feng, Bailan and Ma, Chao},
  journal={arXiv preprint arXiv:2407.12294},
  year={2024},
  url={https://arxiv.org/abs/2407.12294}
}

@article{zhang2025clip,
  title={CLIP prior-guided 3D open-vocabulary occupancy prediction},
  author={Zhang, Zongkai and Gao, Bin and Ye, Jingrui and Jin, Huan and Jiang, Lihui and Yang, Wenming},
  journal={Pattern Recognition},
  volume={162},
  pages={111347},
  year={2025},
  publisher={Elsevier}
}

@inproceedings{doruk2026vlmfusion,
  title={VLMFusionOcc3D: VLM Assisted Multi-Modal 3D Semantic Occupancy Prediction},
  author={Doruk, Abdullah Enes and Ates, Hasan F.},
  booktitle={Proceedings of the IEEE/CVF Winter Conference on Applications of Computer Vision (WACV)},
  year={2026},
  url={https://arxiv.org/abs/2603.02609}
}

@inproceedings{wang2023openoccupancy,
  title={OpenOccupancy: A Large Scale Benchmark for Surrounding Semantic Occupancy Perception},
  author={Wang, Xiaofeng and Zhu, Zhengbiao and Xu, Wenbo and Zhang, Yunpeng and Wei, Yi and Chi, Xu and Ye, Yun and Du, Dalong and Lu, Jiwen and Wang, Xingang},
  booktitle={Proceedings of the IEEE/CVF International Conference on Computer Vision (ICCV)},
  pages={17804--17813},
  year={2023},
  url={https://arxiv.org/abs/2303.03991}
}

@inproceedings{zhang2023occformer,
  title={Occformer: Dual-path Transformer for Vision-based 3D Semantic Occupancy Prediction},
  author={Zhang, Yunpeng and Zhu, Zheng and Du, Dalong},
  booktitle={Proceedings of the IEEE/CVF International Conference on Computer Vision (ICCV)},
  year={2023},
  url={https://arxiv.org/abs/2304.05316}
}

@inproceedings{li2023fbocc,
  title={FB-OCC: 3D Occupancy Prediction based on Forward-Backward View Transformation},
  author={Li, Zhiqi and Yu, Zhiding and Austin, David and Fang, Mingsheng and Lan, Shiyi and Kautz, Jan and {\'A}lvarez, Jos{\'e} Manuel},
  booktitle={Proceedings of the IEEE/CVF Conference on Computer Vision and Pattern Recognition (CVPR)},
  year={2023},
  url={https://arxiv.org/abs/2307.01492}
}

@inproceedings{pan2024renderocc,
  title={RenderOcc: Vision-Centric 3D Occupancy Prediction with 2D Rendering Supervision},
  author={Pan, Mingjie and Liu, Jiaming and Zhang, Renrui and Huang, Peixiang and Li, Xiaoqi and Liu, Li and Zhang, Shanghang},
  booktitle={Proceedings of the IEEE/CVF Conference on Computer Vision and Pattern Recognition (CVPR)},
  year={2024},
  url={https://arxiv.org/abs/2309.09502}
}

@inproceedings{huang2024selfocc,
  title={SelfOcc: Self-Supervised Vision-Based 3D Occupancy Prediction},
  author={Huang, Yuanhui and Zheng, Wenzhao and Zhang, Borui and Zhou, Jie and Lu, Jiwen},
  booktitle={Proceedings of the IEEE/CVF Conference on Computer Vision and Pattern Recognition (CVPR)},
  pages={19946--19956},
  year={2024},
  url={https://arxiv.org/abs/2311.12754}
}

@article{zhang2023occnerf,
  title={OccNeRF: Advancing 3D Occupancy Prediction in LiDAR-Free Environments},
  author={Zhang, Chubin and Yan, Juncheng and Wei, Yi and Li, Jiaxin and Liu, Li and Tang, Yansong and Duan, Yueqi and Lu, Jiwen},
  journal={IEEE Transactions on Image Processing},
  year={2025},
  volume={34},
  pages={3096--3107},
  url={https://arxiv.org/abs/2312.09243}
}

@inproceedings{wang2024panoocc,
  title={PanoOcc: Unified Occupancy Representation for Camera-based 3D Panoptic Segmentation},
  author={Wang, Yu-Quan and Chen, Yuntao and Liao, Xingyu and Fan, Lue and Zhang, Zhaoxiang},
  booktitle={Proceedings of the IEEE/CVF Conference on Computer Vision and Pattern Recognition (CVPR)},
  year={2024},
  url={https://arxiv.org/abs/2306.10013}
}

@article{feng2024vipocc,
  title={ViPOcc: Leveraging Visual Priors from Vision Foundation Models for Single-View 3D Occupancy Prediction},
  author={Feng, Yi and Han, Yu and Zhang, Xijin and Li, Tanghui and Zhang, Yanting and Fan, Rui},
  journal={arXiv preprint arXiv:2412.11210},
  year={2024},
  url={https://arxiv.org/abs/2412.11210}
}

@inproceedings{li2022deep,
  title={Deep Hierarchical Semantic Segmentation},
  author={Li, Liulei and Zhou, Tianfei and Wang, Wenguan and Li, Jianwu and Yang, Yi},
  booktitle={Proceedings of the IEEE/CVF Conference on Computer Vision and Pattern Recognition (CVPR)},
  pages={1246--1257},
  year={2022},
  url={https://arxiv.org/abs/2203.14335}
}

@article{zuo2026quadricformer,
  title={Quadricformer: Scene as superquadrics for 3d semantic occupancy prediction},
  author={Zuo, Sicheng and Zheng, Wenzhao and Han, Xiaoyong and Yang, Longchao and Lu, Jiwen and others},
  journal={Advances in Neural Information Processing Systems},
  volume={38},
  pages={47779--47801},
  year={2026},
  url={https://arxiv.org/abs/2506.10977}
}

@inproceedings{huang2025gaussianformer2,
  title={Gaussianformer-2: Probabilistic gaussian superposition for efficient 3d occupancy prediction},
  author={Huang, Yuanhui and Thammatadatrakoon, Amonnut and Zheng, Wenzhao and Zhang, Yunpeng and Du, Dalong and Lu, Jiwen},
  booktitle={Proceedings of the computer vision and pattern recognition conference},
  pages={27477--27486},
  year={2025},
  url={https://arxiv.org/abs/2412.04384}
}

@inproceedings{murez2020atlas,
  title={Atlas: End-to-end 3d scene reconstruction from posed images},
  author={Murez, Zak and Van As, Tarrence and Bartolozzi, James and Sinha, Ayan and Badrinarayanan, Vijay and Rabinovich, Andrew},
  booktitle={European conference on computer vision},
  pages={414--431},
  year={2020},
  url={https://arxiv.org/abs/2003.10432}
}

@inproceedings{Leng2025OccupancyLW,
  title={Occupancy Learning with Spatiotemporal Memory},
  author={Leng, Ziyang and Yang, Jiawei and Yi, Wenlong and Zhou, Bolei},
  booktitle={Proceedings of the IEEE/CVF International Conference on Computer Vision (ICCV)},
  pages={26569--26578},
  year={2025},
  url={https://arxiv.org/abs/2508.04705}
}

@article{Zhu2026DrOccDA,
  title={{Dr.Occ}: Depth- and Region-Guided {3D} Occupancy from Surround-View Cameras for Autonomous Driving},
  author={Zhu, Xubo and Zhang, Haoyang and He, Fei and Wu, Rui and Shan, Yanhu and Yang, Wen and Yu, Huai},
  journal={arXiv preprint arXiv:2603.01007},
  year={2026},
  url={https://arxiv.org/abs/2603.01007}
}

@article{Kim2026ClassDistributionGA,
  title={Class-Distribution Guided Active Learning for {3D} Occupancy Prediction in Autonomous Driving},
  author={Kim, Wonjune and Lee, In-Jae and Hwang, Sihwan and Kim, Sanmin and Kum, Dongsuk},
  journal={IEEE Robotics and Automation Letters},
  volume={11},
  pages={6999--7006},
  year={2026},
  url={https://arxiv.org/abs/2603.27294}
}

@article{Mattamala2024WildVN,
  title={Wild visual navigation: fast traversability learning via pre-trained models and online self-supervision},
  author={Mattamala, Mat{\'\i}as and Frey, Jonas and Libera, Piotr and Chebrolu, Nived and Martius, Georg and Cadena, C{\'{e}}sar and Hutter, Marco and Fallon, Maurice F.},
  journal={Autonomous Robots},
  volume={49},
  year={2024},
  url={https://arxiv.org/abs/2404.07110}
}
\end{document}